\documentclass[review]{elsarticle}
\usepackage[legalpaper, margin=1in]{geometry}
\usepackage{graphicx}
\usepackage{amssymb}
\usepackage{lineno}
\usepackage{lineno,hyperref}
\usepackage{amsmath}
\modulolinenumbers[5]
\usepackage{subfig}
\usepackage[figurename=Fig.]{caption}
\journal{Ocean Engineering}

\begin{document}
\begin{frontmatter}


\title{Machine Learning for Naval Architecture, Ocean and Marine Engineering}



\author{J.P. Panda} 
\address{Department of Mechanical Engineering,
DIT University, Dehradun, Uttarakhand,  India}

\fntext[myfootnote]{corresponding author: jppanda.iit@gmail.com }

\begin{abstract}
Machine Learning (ML) based algorithms have found significant impact in many fields of engineering and sciences, where datasets are available from experiments and high fidelity numerical simulations. Those datasets are generally utilized in a machine learning model to extract information about the underlying physics and derive functional relationships mapping input variables to target quantities of interest. Commonplace machine learning algorithms utilized in Scientific Machine Learning (SciML) include neural networks, regression trees, random forests, support vector machines, etc. The focus of this article is to review the applications of ML in naval architecture, ocean and marine engineering problems; and identify priority directions of research. We discuss the applications of machine learning algorithms for different problems such as wave height prediction, calculation of wind loads on ships, damage detection of offshore platforms, calculation of ship added resistance and various other applications in coastal and marine environments. The details of the data-sets including the source of data-sets utilized in the ML model development are included. The features used as the inputs to the ML models are presented in detail and finally the methods employed in optimization of the ML models were also discussed. Based on this comprehensive analysis we point out future directions of research that may be fruitful for the application of ML to ocean and marine engineering problems.     
          
\end{abstract}

\end{frontmatter}


\section{Introduction}
\label{S:1}

In the fields of naval architecture, ocean and marine engineering  large amounts of data are generated from sources such as ocean wave and current measurements, sea floor mapping by AUVs, ship manoeuvring and recorded wind speeds interacting with the ships at different locations across the globe, etc. This data potentially represents latent knowledge that can advance our understanding of and our solutions for these fields. But the volume and diversity of this data limits manual analysis by human domain experts. Using machine learning based techniques researchers can utilize, interpret, visualize and analyze this data. This enables the generation of data driven surrogate models for these physical phenomena. These predictive models can be utilized for estimation of wave height and ship parameters such as container capacity and added mass coefficient, etc. 

Machine learning (ML) is a branch of Artificial Intelligence (AI) that focuses on enabling computers to infer models from data and constraints. The various steps involved in developing a ML model are data preparation, feature engineering, data modelling, and model evaluation. Data preparation involves collection of raw data, data cleaning (deals with the missing values and removing outliers) and formatting the raw data that can be incorporated in a machine learning model. Feature Engineering is the process of converting the raw data into physics based features, which can be correlated with a quantity of interest in a particular field of engineering. One example of quantity of interest in the field of Marine Engineering is the drag coefficient of an autonomous underwater vehicle (AUV), which has relations with size of the AUV, velocity of the flow, upcoming turbulence intensity and various other environmental factors. Feature Engineering provides adequate data for a machine learning model that can enhance the performance of the model. Next step of the of the machine learning is data modelling, that is splitting data into training and testing sets. In the training process both the inputs and the quantity of interests(outputs) are provided to the model. The machine learning algorithm maps the input variables with outputs and give a target function which can be used to predict the unknown parameters for other set of input variables(testing data).   

There are mainly four different types of machine learning techniques. Those are supervised, unsupervised, semi-supervised and reinforcement learning. In supervised machine learning the outputs for a given set of inputs are used for training the model and once the model is trained, it is used for prediction. In unsupervised learning the outputs for given inputs are unknown. The semi-supervised learning is in between supervised  and unsupervised learning where some samples may have training labels and others may not. In reinforcement learning the machine is exposed to the environment, where it learns by optimizing its reward. Among all the machine learning techniques, the supervised learning methods are most widely adopted in the engineering community. In supervised learning labelled data used for for training and problems of classification and regression are solved. Popular supervised machine learning algorithms are Linear Regression, Support Vector Machines (SVM), Neural Networks, Decision Trees, Naive Bayes, etc.

In this article we provide a detailed review of application of ML algorithms in the naval architecture, ocean and marine Engineering and grouped those into following categories: wave forecasting, AUV operation and control, applications in ship research, design and reliability analysis of breakwaters, detection of damaged mooring lines, applications in propeller research, damage detection of offshore platforms and few other miscellaneous applications like beach classification, condition monitoring of marine machinery system, Performance assessment of shipping operations, autonomous ship hull corrosion cleaning system, wave energy forecasting, prediction of wind and wave induced load effects on floating suspension bridges and tidal current prediction. Based on this comprehensive analysis we point out future directions of research that may be fruitful for the application of ML to coastal and marine problems.

\section{Machine learning basics}
Machine learning is the process of finding the associations between inputs and outputs and parameters of a system using limited data. The learning process can be summarized as follows \citep{cherkassky2007learning}:    
\begin{equation}
\begin{split}
R(w)=\int L[y, \phi(x,y,w)]p(x,y)dxdy
\end{split}
\end{equation}
In the above equation, data x and y are the samples form the probability distribution $p$, the structure of the ML model is defined by $\phi(x,y,w)$, and $w$ are the parameters. The learning objectives are balanced by the loss function $L$. There are three broad categories of ML algorithms, those are  supervised, unsupervised, and semi-supervised.
\subsection{Supervised Learning}
In supervised learning, correct information is available to the ML algorithm. The data utilized for the development of the ML model are labeled data, where labels are available corresponding to the output. The unknown parameters of the ML model are determined through minimization of the cost function. The supervised learning correspond to various regression and interpolation methods.The loss function of the ML model can be simply defined as:
\begin{equation}
\begin{split}
L[y, \phi(x,y,w)]=|x||y|b
\end{split}
\end{equation}
 \begin{figure}
\centering
\includegraphics[height=5.4cm]{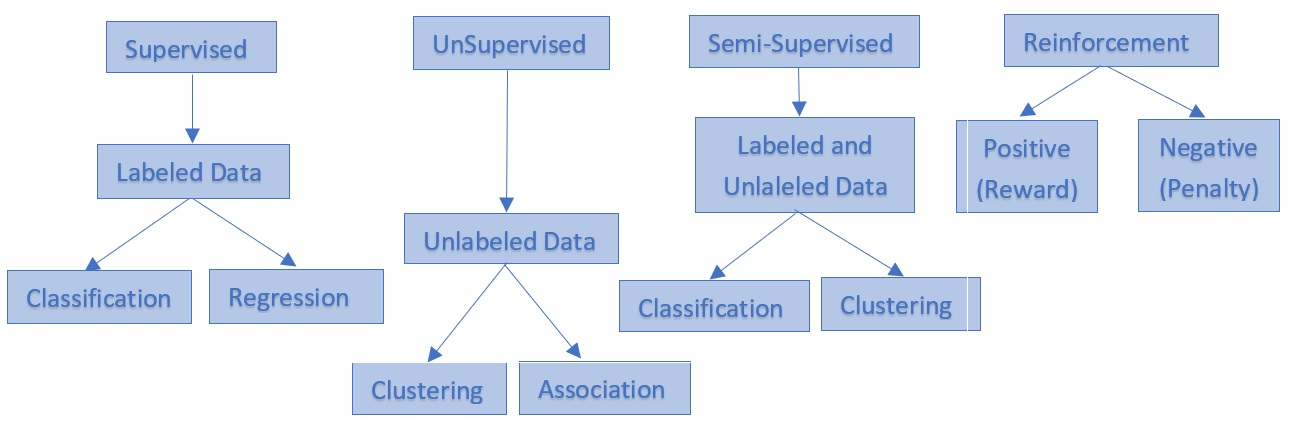}
\caption{Classification of ML algorithms as supervised, semi-supervised and unsupervised and reinforcement learning.} \label{fig:1}
\end{figure}

\begin{figure}
\centering
\includegraphics[height=6.2cm]{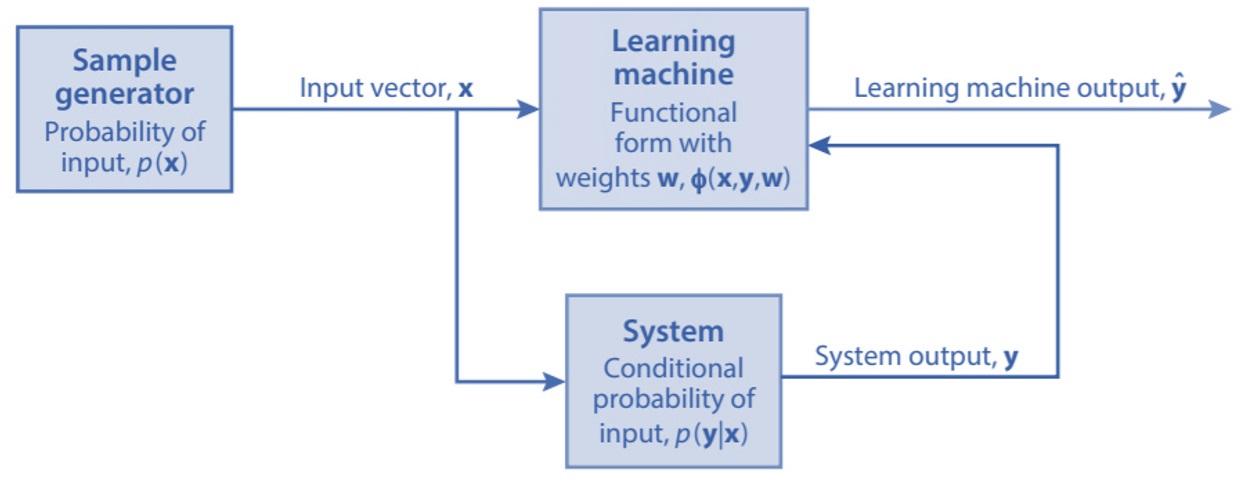}
\caption{The learning problem} \label{fig:lp}
\end{figure}
\subsection{Unsupervised Learning}
In unsupervised learning, the features are extracted from the data by specification of certain criteria and supervision and ground truth levels are not required. The problems involved in unsupervised learning are clustering, quantization and dimensionality reduction. The dimensionality reduction involve proper orthogonal decomposition, autoencoders and principal component analysis. In clustering, similar groups in data can be identified. The most common ML algorithm used in clustering of data is the k-means clustering \citep{hartigan1979algorithm}.
\subsection{Semisupervised Learning}
In semisupervised learning, the ML algorithm is partially supervised, either with corrective information from the environment or with limited labeled training data. In semi-supervised learning two alorithms are mainly used, those are generative adversarial networks \citep{creswell2018generative} and Reinforcement learning \citep{sutton2018reinforcement}.

\section{Popular Machine learning algorithms}
\label{S:2}
\subsection{Artificial Neural Network}
\begin{figure}
\centering
\includegraphics[height=8cm]{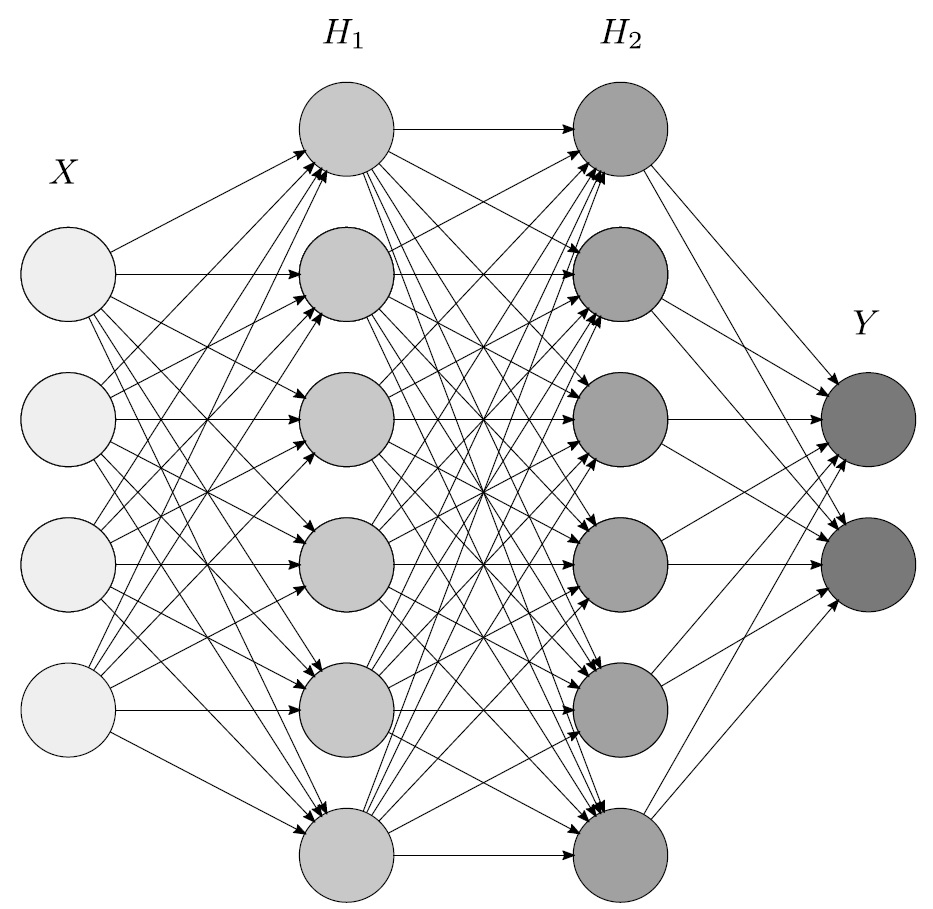}
\includegraphics[height=6cm]{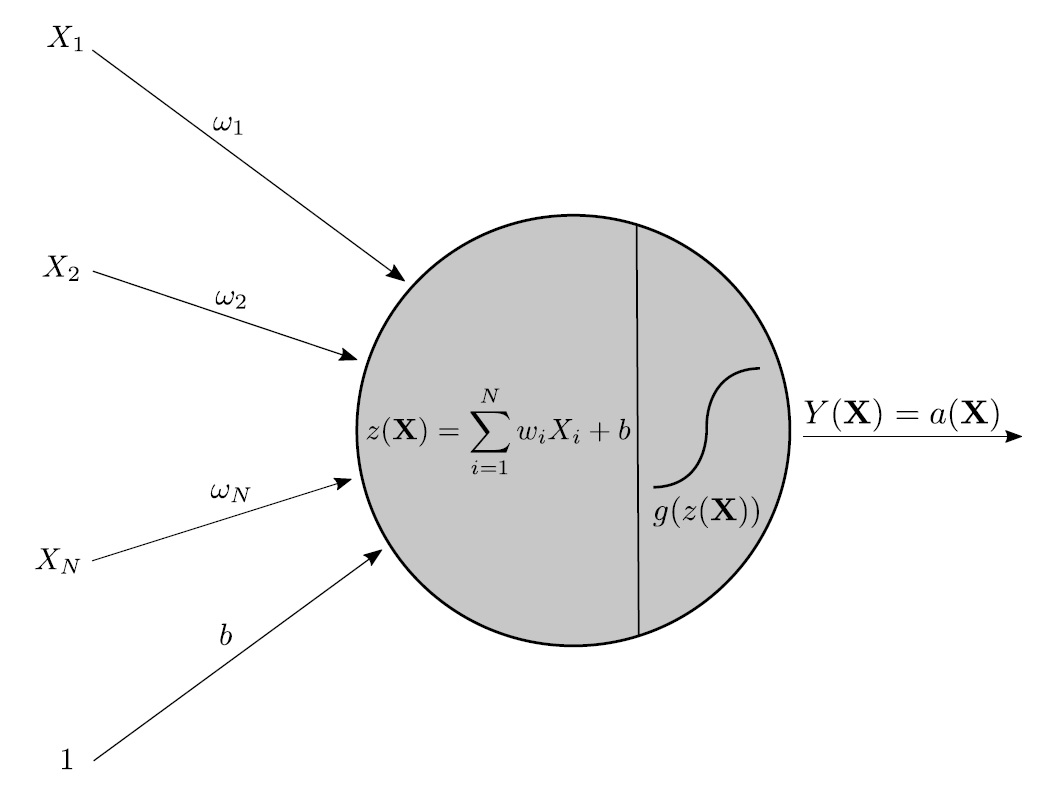}
\caption{Structure of a neural network} \label{fig:ann}
\end{figure}

Artificial Neural Networks(ANN) (fig.\ref{fig:ann}a) are the machine learning systems inspired from the biological neural networks. The biological neural networks(BNN) are the circuits that carry out a specific task when activated. These are population of neurons interconnected by synapses. Similar to BNNs, ANNs have artificial neurons(fig.\ref{fig:ann}b). The most widely used ANN is a multilayer perception(MLP), which has more than one hidden layers and has applications in both regression and classification problems. The MLP has input layer, hidden layers and output layer. The MLP correlates inputs to the outputs. While passing through the hidden layers the inputs are multiplied by weights. Each neuron in the MLP has a function y that correlates the input. The ultimate aim is to reduce the error at the output by optimizing the weights. In every layer of the neural network, the neuron response is given by an activation function, and a cost is given by a biased weighted sum. For two consecutive layers $[k-1,k]$, the neural network operation can be expressed mathematically as follows:
\begin{equation}
\begin{split}
y_j=f_j(\sum_{i=1}^{n}w_{ij}x_{i}+b_{j}),i\in[0,n]\wedge i\in[0,m] 
\end{split}
\end{equation}
where $n$ and $m$ are number of layers in $k-1$ and $k$th layers respectively. For neuron $j$, $y_{j}$ is the output and $x_{i}$ is the input signal from neuron $i$, $b_j$ is the bias of the neuron and $w_{ij}$ is the weight associated with connections $i$ and $j$.

The MLP can learn from data and optimize the weights. The learning task in DNNs is obtained by updating the weights and minimizing the output error. The DNNs employ back-propagation algorithm in the training process and it is good for both speed and performance. There are several optimization techniques by which the weights can be calculated. Those are gradient descent, Quasi- Newton, Stochastic Gradient Descent or the Adaptive moment Estimation etc.
\subsection{Convolutional Neural Network}
\begin{figure}
\centering
\includegraphics[height=4.2cm]{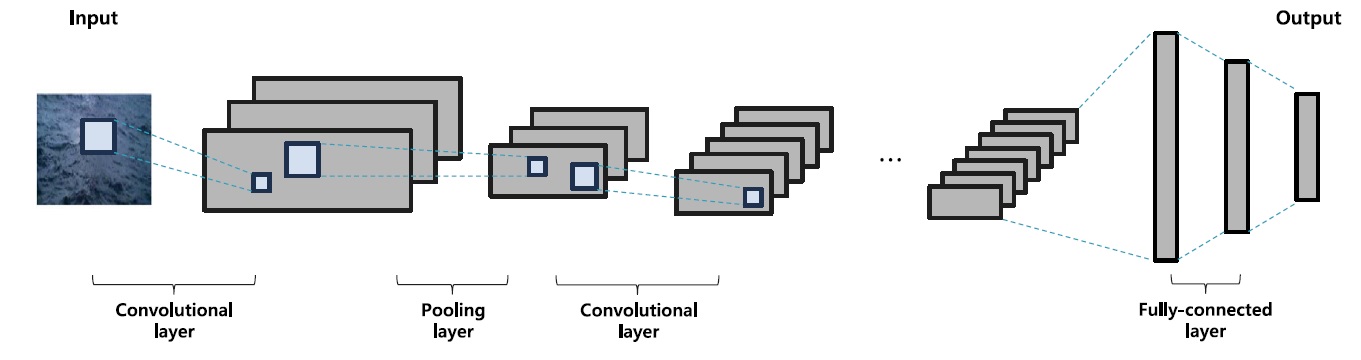}\\
\caption{Architecture of CNN} \label{fig:2}
\end{figure}

Convolutional neural network (CNN) is type of neural network mainly used in image processing. The CNN has mainly three layers, those are convolutional layer, pooling layer and a fully connected layer. The schematic of CNN is shown in fig.\ref{fig:2}. The job of convolutional layer is to extract latent features form an image by applying convolutional filters to the input. A set of learnable filters are the parameters of CNN connected only to a local region in the input volume spatially, but to the full depth. The detailed methodology involved in CNN can be presented as follows:
\begin{equation}
\begin{split}
x^l_{i,j}=g(x^{l-1}*W^l)_{i,j}=g(\sum_m\sum_n x_{{i+m},{j+n}}^{l-1}w^l_{m,n})
\end{split}
\end{equation}
in above equation, $x^l_{i,j}$ is (i,j)th value of $l$th layer and $w^l_{m,n}$ is ($m,n)$th weight of convolution filter in the $l$th layer. g is the non-linear activation function. Suppose, the $(l-1)$th layer has dimensions of W(width), H(height) and C(Channel) and the $l$th layer convolutional filter dimensions F(width), F(height) and C(Channel), then the  $l$th layer can be derived by applying the nonlinear activation function g to the convolution operation. The function of nonlinear activation function is to model the nonlinear relationship the subsequent layers.

\subsection{Recurrent neural network}
Recurrent neural networks (RNN) are used for processing sequential data. RNNs can also process data with much longer sequences and sequences with variable length. RNNs can be designed in different methods as discussed in \cite{goodfellow2016deep}: a) a RNN that generate an output at every time step and recurrent connections between the hidden units, b) a RNN that generate an output at every time step and recurrent connectivity only from the output at one time step to the hidden units at the next time step and c) RNNs with recurrent connectivity among the hidden units that reads entire sequence and produce a single output. 
\subsubsection{Long short-term memory network}
Long short-term memory network(LSTM) are the gated RNNs. These networks are based on the concept of generating paths through time that neither vanish nor explode. In the LSTM, the gradients can flow for long duration through the self loops. By making weight of the self loop gated the time scale of integration can be changed dynamically. The LSTM has successful applications in handwriting recognition, speech recognition, handwriting generation and machine translation.

\begin{figure}
\centering
\includegraphics[height=10cm]{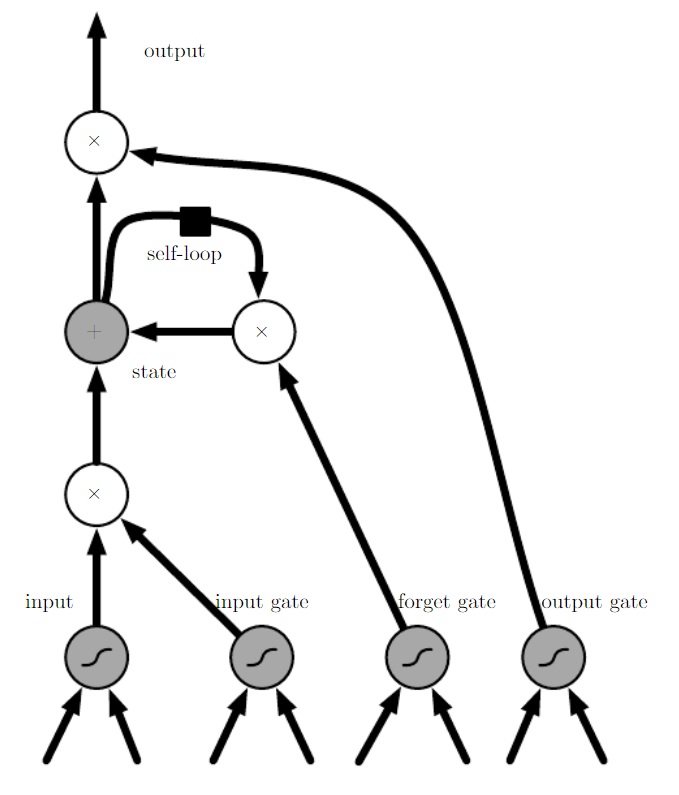}
\caption{Block diagram of the LSTM network cell, The cells are connected recurrently to each other \citep{goodfellow2016deep}. \label{fig:lstm}}
\end{figure}

\subsection{Random Forest}
\begin{figure}
\centering
\includegraphics[height=6.2cm]{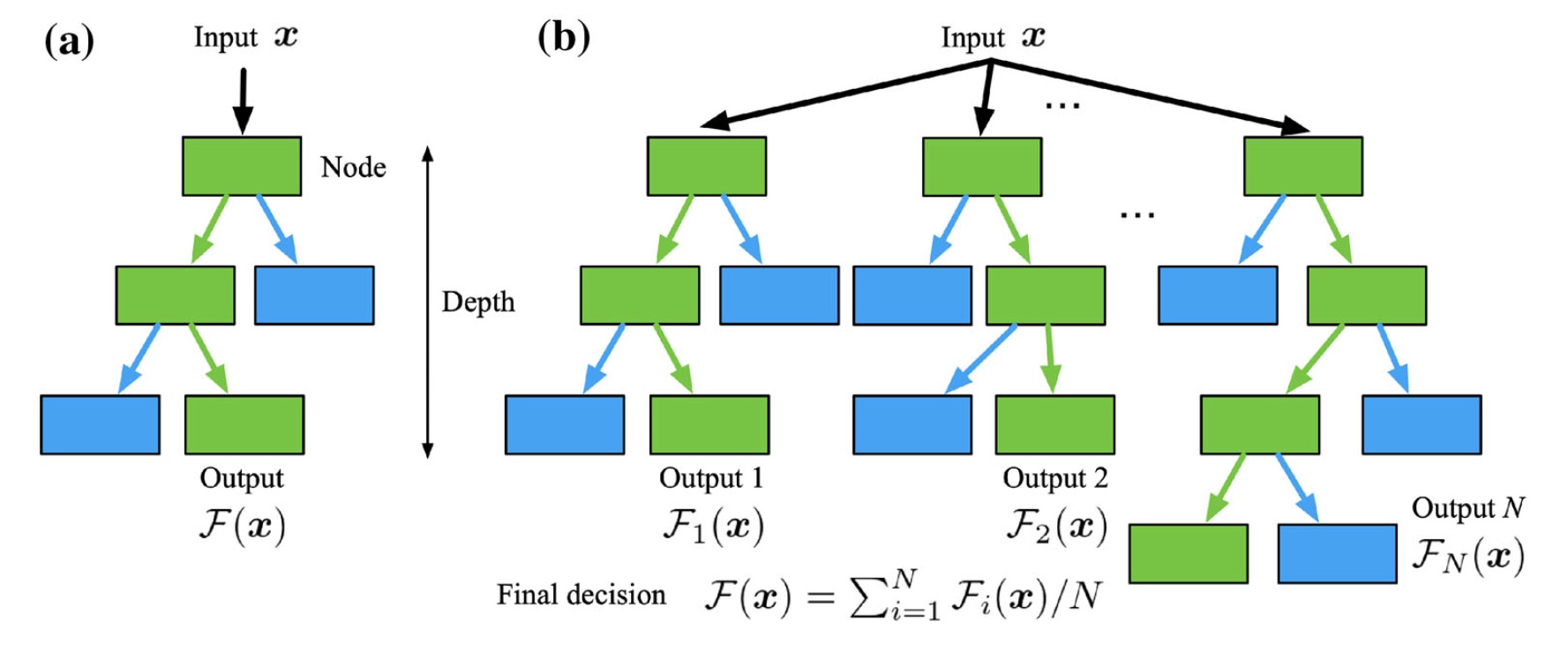}\\
\caption{Architecture of of a random forest, a)a single decision tree b)a random forest} \label{fig:3}
\end{figure}

The basic building block of a random forest is a decision tree. The structure of the decision tree is shown in fig. \ref{fig:3}a. The boxes in the decision tree represent group of features and data. The decision tree seeks if/then/else rules for obtaining the desired output.     Random forest(RF) regression combines performance of multiple decision trees for predicting a output variable. It is a assemble learning technique, works on the concept of bagging method. In RF regression trees are constructed using a subset of random sample drawn with replacement from training data. The RF algorithm has following steps:
a) Drawing boot strap samples for number of trees from original data
b) Growing unpruned regression trees for each boot strap sample and at each node randomly sample the number of predictors and choose the best split among those variables.
c) Prediction of output for test data by the averages of the predictions from all the trees.
 
\subsection{Support Vector Machine}
Support vector machines are based on the principles of structural risk minimization and are used for recognizing the subtle patterns in complex datasets. This machine learning methodology was initially developed for solving the classification problems, later these were extended towards solving regression problems. The support vector regression employs kernel functions to map the initial data into high dimensional space such that the nonlinear patterns can be converted
into a linear problem. The performance of SVM regressor is largely dependent on the choice of kernel function. There are 4 main types of kernel functions are available, those are Linear, Polynomial, Sigmoid and Radial Basis Function. The hyper-parameters of the SVM regressor must be chosen carefully for efficient performance of the model. Inappropriate choice of hyper parameters would lead to under/ over fitting. The two main hyper parameters of the SVM regression model are $\epsilon$-insensitive zone and regularization parameter C. Any deviation from $\epsilon$ can be calculated from the penalization of the regularization parameter C. The penalty becomes more important when the values of C is higher and SVR fits the data. The penalty becomes negligible, when the value of C is small and SVR gets flat. For higher value of $\epsilon$ SVR becomes flat and for smaller value of $\epsilon$, SVR fits data.            
\section{Application of Machine learning in the marine environment} \label{S:3}
\subsection{Wave forecasting}
Waves are generated by complex interaction of wind with the ocean and the process by which ocean waves are generated is not fully understood and is extremely uncertain and complex. The information of wave heights at different locations is essential for operation related activities in the ocean. Most of the works related to operation activities in the ocean are carried out with wave heights estimated over a period of some hours or days. Traditionally deterministic models are used for prediction of height of waves and wave periods   \citep{sandhya2014wave}. The ocean waves can both be forecasted and hindcasted using different physics based approaches. Waves can be forecasted using meterological conditions and can be hindcasted with different meterological charts. In wave forecasting, differential equations for wind-wave relationship and wave energy are solved numerically.
The methods using differential equations generally predict wave heights for a period of 6-72 hours. The cost of numerical simulations using differential equations is very high and the simulations are time consuming and the numerical predictions are always associated with uncertainties. The uncertainties appears in the prediction results because of the approximations utilized in the model development.             

Because of the limitations in the traditional methods of wave height predictions and the rapid development of ML methods, now researchers have started utilizing various ML algorithms \citep{mafi2017forecasting,etemad2009comparison,deka2012discrete,wang2018bp,law2020deterministic,oliveira2020impact,deshmukh2016neural} such as neural networks \citep{deo2001neural,deo1998real,sahoo2019prediction}, recurrent neural networks \citep{mandal2006ocean} and  random forests for prediction of wave heights.
\begin{figure}
\centering
\includegraphics[height=5cm]{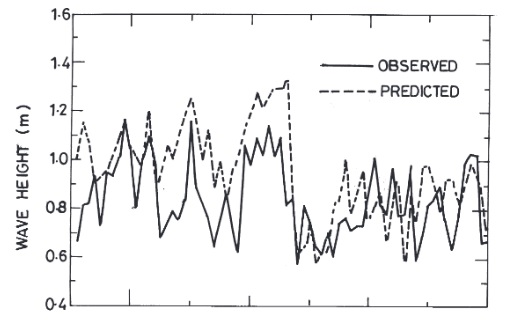}
\includegraphics[height=5cm]{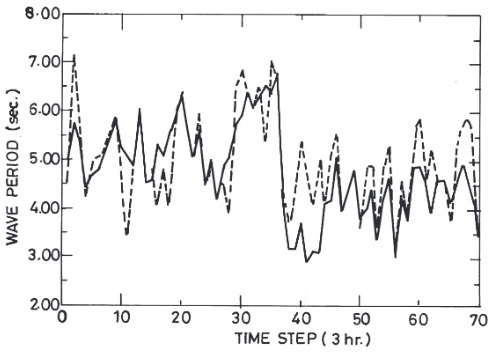}
\caption{Neural network prediction of wave heights and periods \citep{deo2001neural}.} \label{fig:deo}
\end{figure}
\cite{deo2001neural} have used a simple 3 layered feed forward neural network to predict the significant wave height and wave period. They considered the wind speed as input to the neural network for March 1988 to July 1988 and further from December 1988 to May 1989. The location of the data collection point was off Karwar in India. The ANN prediction of wave height is shown in fig. \ref{fig:deo}. \cite{tsai2009wave} have used ANN to predict the significant wave height, significant wave period, maximum wave height and spectral peakedness parameter. The input parameters of the neural network are       : average of the highest one third pressure wave heights, corresponding pressure wave period, average pressure zero-crossing period, maximum pressure wave height , average pressure wave height, root mean square pressure wave height, average of one-tenth highest pressure wave height, successive pressure wave height correlation and pressure spectral peakedness parameter. The detailed definitions of the above parameters are available in \citep{tsai2009wave}. The ANN model was trained using data obtained from stations ranged from 11 to 41 m. The predicted results are compared with the results obtained from linear theory. The ANN model predicted better results for water depths in between 20 to 41 meter in comparison to the linear theory.
\cite{rao2005hindcasting} used neural networks to estimate wave parameters from cyclone generated wind fields. They considered 11 cyclones, which crossed the southern east coast of India in their studies. The inputs to the neural network are difference between central and peripheral pressure, radius of maximum wind and the speed of forward motion of
cyclone. The outputs of the neural network are wave heights and periods. They considered the feed forward neural network with back propagation algorithm in their modelling. The NN model predictions are contrasted against other established wave hindcasting models and the observed very good correlation between NN and physics based model results.

\begin{figure}
\centering
\includegraphics[height=8cm]{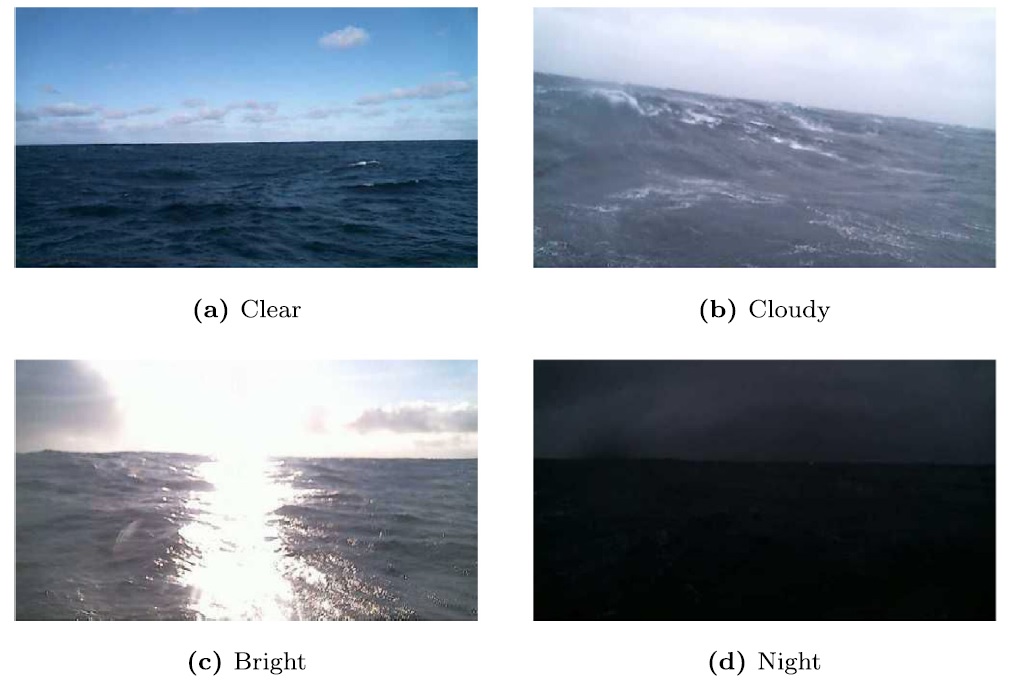}
\caption{Raw ocean images \citep{choi2020real}.} \label{fig:ocean_i}
\end{figure}

\begin{figure}
\centering
\includegraphics[height=8cm]{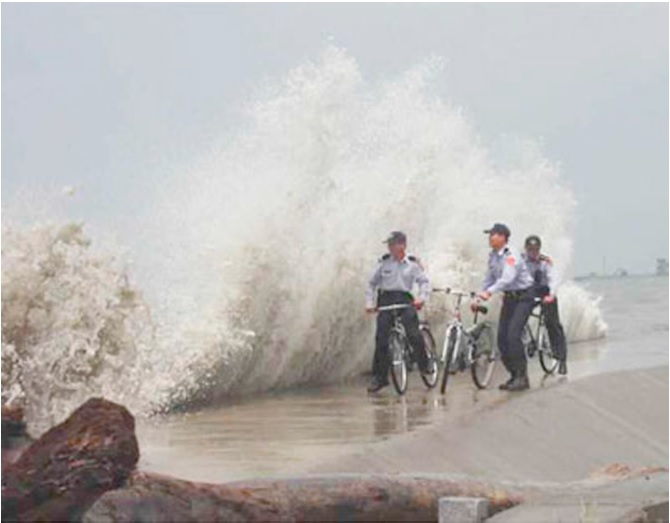}\\
\caption{An actual picture of a coastal freak wave \citep{doong2018development}.} \label{fig:4}
\end{figure}

\cite{oh2018real} used wavelet and neural network hybrid models for real-time forecasting of wave heights. They developed the hybrid model by combining empirical orthogonal function analysis and wavelet analysis with the neural network and used wave height data at different locations and meteorological data in the surrounding are for training the ANN model. Their developed model was useful for prediction of wave heights where past wave height and meteorological data are available. Doong et al. used ANN based models to predict the the occurrence of coastal freak waves. An actual picture of a coastal freak wave is shown in fig. \ref{fig:4}. These waves are generated by interaction of waves with various coastal structures such as rocks. They used seven parameters(significant wave
height, peak period, wind speed, wave groupiness factor, Benjamin Feir Index (BFI), kurtosis, and wind-wave direction misalignment) for training their model. They used a single hidden layer neural network with back propagation algorithm. The field data used in their study were collected from the Longdong Data Buoy(Central weather bureau of Taiwan). The buoy was at a location of 1 km off the Longdong coast, where the water depth was 23m. \cite{choi2020real} used deep neural networks to estimate significant wave height using raw ocean images\ref{fig:ocean_i}. A CNN based classification model was constructed with four CNN structures. Their method of wave height prediction had two steps: a) the best CNN model for ocean image processing was found among VGGNet, Inception.v3, ResNet and DenseNet. b) The model performance was improved using transfer learning and structure modification. It was observed that the VGGNet and ResNet based model with transfer learning and various feature extractors yielded good performance in significant wave height modeling. 

\begin{figure}
\centering
\includegraphics[height=9cm]{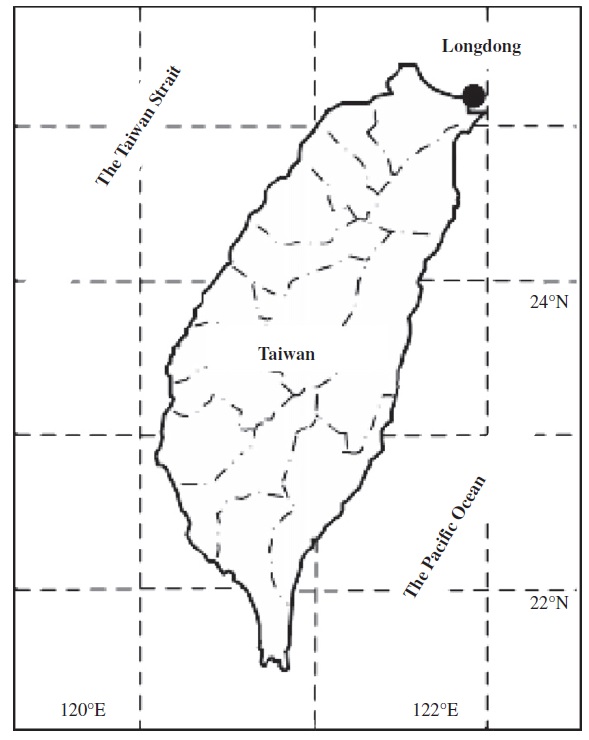}
\caption{Locations of Longdong harbour, Taiwan \citep{rajasekaran2008support}.} \label{fig:taiwan}
\end{figure}

Traditionally strom surges are predicted using fluid dynamics methods(finite difference method) that utilizes large number of equations and the simulations are time consuming. \cite{rajasekaran2008support} applied support vector regression methodology to forecast typhoon surge. The typhoon surge must be accurately predicted to avoid property loss. For the development of ML model they used the input data are pressure, wind velocity, wind direction and estimated astronomical tide and the out put is the storm surge level. The ML model developed with support vector regression was verified using original data collected at the Longdong station at Taiwan for the Aere typhoon. The location of the Longdong harbour is shown in fig. \ref{fig:taiwan}. 
\cite{malekmohamadi2011evaluating} evaluated the efficacy of support vector machine, bayesian networks and artificial neural networks in predicting the wave height. The data required for training the models were collected at a buoy station at Lake Superior. They noticed ANN prediction of the wave height matched well with the observational data and are better than prediction of the other ML models.        

\subsection{AUV operation and control}
An AUV(autonomous underwater vehicle) is a self propelled unmanned vehicle, that can conduct various activities in the deepest corners of Ocean or near the free surface \citep{tyagi2006calculation}. These don't involve any human supervision. Typical applications of AUV includes, sea floor mapping for construction of offshore structures, characterization physical chemical and biological properties of ocean, Oceanographic applications and many more. Since there is minimal human supervision for AUV operation, design of accurate control system of AUV plays major role in the design of AUV. These vehicles have are robotic devices with own propulsion system for navigation and have onboard computer for decision making \citep{SAHOO2019145}. 

\begin{figure}
\centering
\includegraphics[height=5cm]{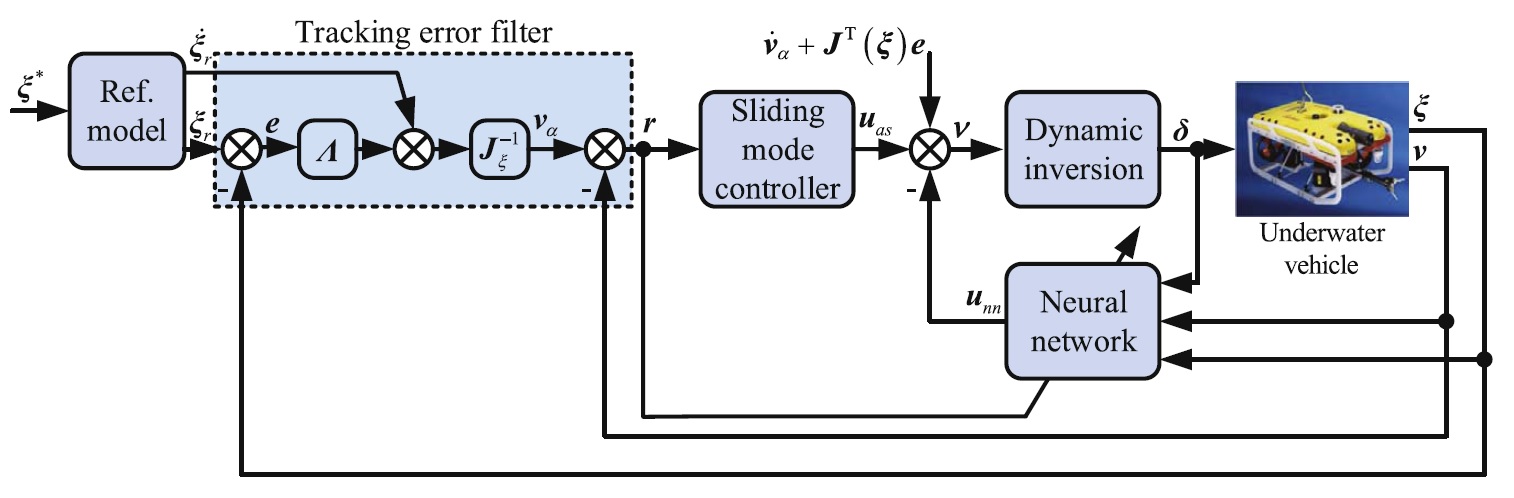}\\
\caption{Schematic of the DI-SMANNC for hybrid visual servoing of an underwater vehicle \citep{gao2017sliding}.} \label{fig:5}
\end{figure}

Machine learning techniques has various application in the design of control system of the AUV.  \cite{zhang2020neural} used neural networks for modelling an adaptive trajectory tracking control scheme for under actuated autonomous underwater. The AUV was subjected to unknown asymmetrical actuator saturation and unknown dynamics. The AUV kinematic controller was designed by using neural network compensation and adaptive estimation techniques. The neural network was used to approximate the complex AUV hydrodynamics and differential of desired tracking velocities. The stability of the NN model was tested against Lyapunov theory and backstepping technique.  \cite{zhang2018master} proposed a novel bilateral adaptive control scheme for achieving position and force coordination performance
of underwater manipulator teleoperation system under model uncertainty and external disturbance. A new nonlinear model reference adaptive impedance controller with bound-gain-forgetting (BGF) composite
adaptive law is designed for the master manipulator force tracking of the slave manipulator. They have used a radial basis function neural network   
for local approximation of slave manipulator's position tracking. The RBFNN based on Ge-Lee (GL) matrix is adopted to directly approximate each element of the slave manipulator dynamic, and the robust term with a proper update law is designed to suppress the error between the
estimate model and the real model, and the external disturbance. \cite{gao2017sliding} proposed a hybrid visual servo(HVS) controller for underwater vehicles using a dynamic inversion based sliding mode adaptive neural network control. The method was developed for tracking the HVS reference trajectory generated from a constant target pose. The dynamic uncertainties were compensated using a single layer feed forward neural network, with an adaptive sliding mode controller. The control system proposed was composed of a sliding model controller combined with a neural network compensator is employed to construct a pseudo control signal required to track a smooth reference trajectory, that is generated by a target pose through a reference model. A dynamic inversion module was also incorporated in the control system to convert the pseudo control into actual thruster control signals by using approximate dynamic model of underwater vehicles. The schematic of the proposed control system architecture is shown in fig.\ref{fig:5}. 
\cite{chu2016observer} proposed an observer based adaptive neural network control approach for a class of remotely operated vehicles 
whose velocity and angular velocity state in the body fixed frame are unmeasured. The thruster control signal was considered as the input to the tracking control system. An adaptive state observer based on local recurrent neural network was proposed to estimate the velocity and angular velocity state online. The adaptive learning method was also used to estimate the scale factor of the thrust model. 
\subsection{Machine learning applied in ship research}
\subsubsection{Estimation of ship parameters}
\cite{ray1996neural} utilized neural networks to predict container capacity of a container ship. They used database of previously built ships to predict the number of containers which can be accommodated on a vessel and to design a container stowage plan. The input to the neural network are the length, breadth, depth, deadweight and speed of the vessel and the output is the container capacity(number of containers).
\cite{margari2018use} used artificial neural networks to predict the resistance hullforms. The hull forms are sixteen in number and those were mainly designed to be used as bulk carriers and tankers by the U.S. Maritime Admisistration. The experimental data for the residual resistance coefficient were used to train and test a series of neural networks. They considered a feed forward multilayered perception in their analysis. The input vector to the neural network consists of length to breadth ratio, the breadth to draft ratio, the block coefficient and the Froude number and the output is the residual resistance coefficient.
\cite{cepowski2020prediction} predicted the added resistance of a ship in regular head waves. Experimental data collected from model test measurements was used to train the neural network model. The predicted added resistance has applications in the preliminary design stage. The function for the added mass coefficient was approximated as:
\begin{equation}
\begin{split}
C_{AW}=f(LBP,B,d,CB,Fn,\lambda/LBP)
\end{split}
\end{equation}
where:
LBP is the length between perpendiculars, B is the breadth, d is the draught,CB is the block coefficient, Fn is the Froude number, $\lambda$ is the wave length and f is the function for the prediction of ship added resistance coefficient.
\cite{zhang2013estimation} used support vector machine for estimation of hydrodynamic coefficients in the mathematical models of ship manoeuvring motion from captive model test results. The towing test and pure sway test were also considered in the model testing. The comparison of the test data and SVM predicted results suggest that, SVM is an effective method  in predicting hydrodynamic coefficients in captive model test. They also noticed that, the SVM model predictions are comparatively poor, for polluted test data and uncertainties in the hydrodynamic models. The efficacy of SVM with polluted test data can be enhanced by treating test data with de-noising approaches. \cite{xu2019hydrodynamic} used a least square support vector machine for estimation of parameters nonlinear manoeuvring model in shallow water.                  
\subsubsection{Automatic Ship docking}
\begin{figure}
\centering
\includegraphics[height=5cm]{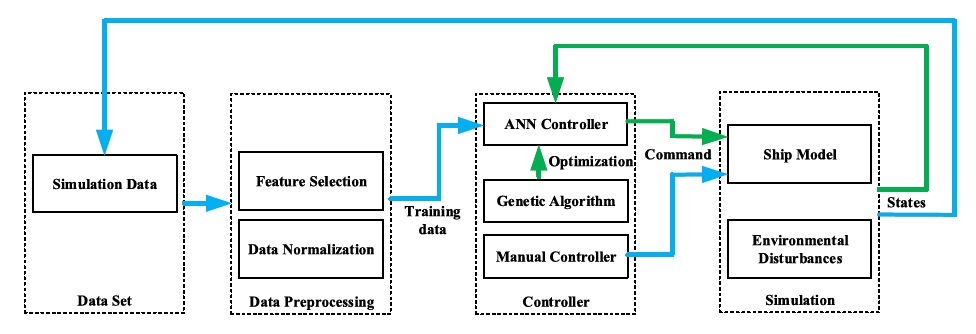}\\
\caption{Schematic of the ANN based control strategy for automated ship docking \citep{shuai2019efficient}.} \label{fig:ann_asd}
\end{figure}

\cite{shuai2019efficient} proposed an ANN based approach for automatic ship docking in presence of environmental disturbances (\ref{fig:ann_asd}). The data required for ANN training was obtained by operating the ship by a skilled captain using a joystick to control ship's rudder and thrust. In the ANN based controller model the inputs are the parameters selected from data analysis(those are state of vessel and environmental information) and the output are the ship's propeller speed and rudder angle. The vessel's state are its heading and position, speed of the vessel, the force of vessel in different degrees of freedom, and the environmental information are force of wind in each degrees of freedom, the wind speed and wind direction.They used gradient descent method to minimize the mean square error between the required output value and the actual output value of the neural network. Based on sensitivity the optimal parameters for the ANN input are chosen as relative distance between ship and dock and the heading angle. The output of the ANN is rudder angle and thruster speed.     
\subsubsection{Ship manoeuvring simulation}
\cite{luo2014manoeuvring} used support vector regression for manoeuvring simulation of catamaran. The implicit models are derived for the manoeuvring motion, instead of the traditional method of calculation hydrodynamic coefficients. For development of the SVM regressor model data obtained from full scale trials were used. The effects of wind and current induced disturbances were also considered in the model development. The inputs to the SVM model were rudder angle, surge speed, yaw rate and sway speed respectively and outputs are derivatives of the surge speed,sway speed and yaw rate respectively. They utilized Gauss function kernel in the SVM to improve the performance of the approximation.
\begin{figure}
\centering
\includegraphics[height=8cm]{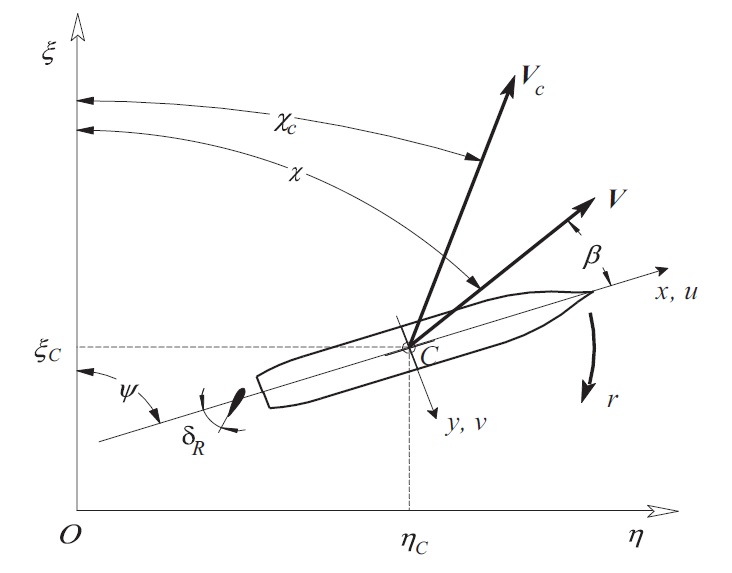}
\caption{Kinematic parameters in ship manoeuvring simulation, detailed definition of these parameters are available in \cite{luo2014manoeuvring}} \label{fig:k_parameter}
\end{figure}
\subsubsection{ship trajectory prediction}
\begin{figure}
\centering
\includegraphics[height=10cm]{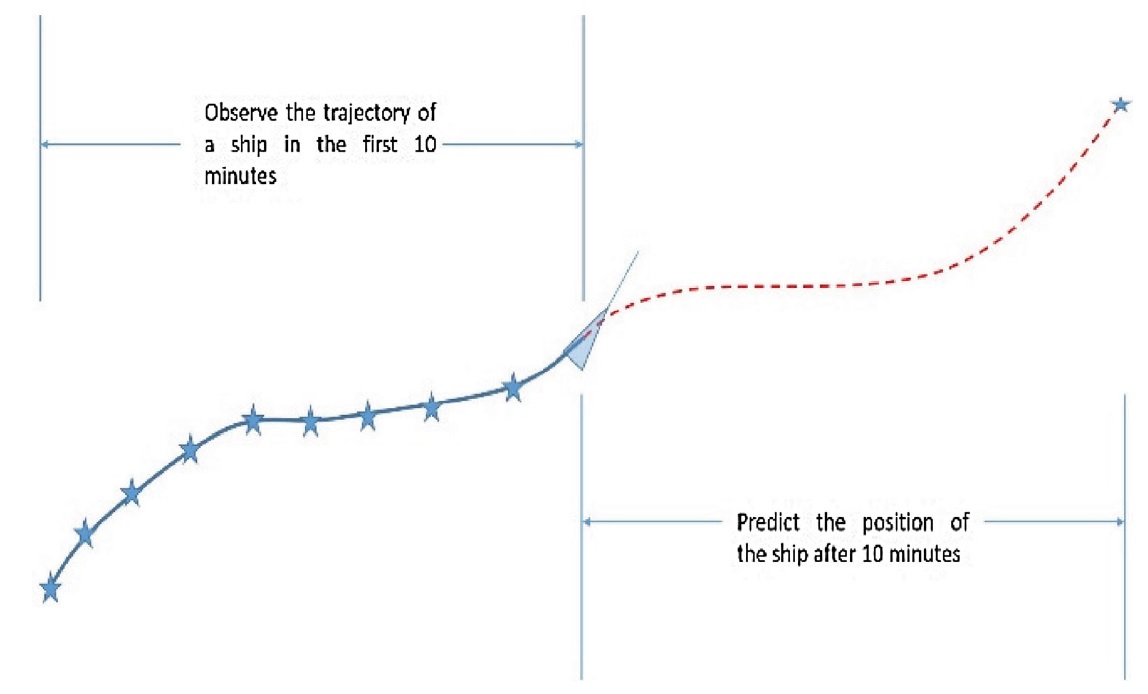}
\caption{Trajectory prediction of the ship \citep{tang2019model}.} \label{fig:trajectory}
\end{figure}

\cite{tang2019model} used Long Short-Term Memory(LSTM) network to model and predict the trajectories of the vessels. The ground truth automatic identification data in the Tianjin port of China was used to train and test the ML model. The inputs to the LSTM model are geographical location, speed, heading and other status information and the output is the status of ship at any future moment. It was observed that, the predicted trajectories are better than the traditional kalman filter model. The predicted trajectory of the ship is shown in fig. \ref{fig:trajectory}. In the figure, the first 10 minute trajectory data was used to train the LSTM model and the trajectory after 10th to 20th minute was predicted by the LSTM model.  
\cite{volkova2021predicting} applied artificial neural networks(ANN) to predict trajectory of the ship. THE AIS data were used for the ANN model development.This method of trajectory planning is mainly useful for river vessels and river-sea vessels, when the vessel is near a hydraulic structure and there is problem in obtaining satellite signals because of interference. The likelihood of collision can be 
decreased with application of ML based trajectory prediction models. 
\subsubsection{Calculation of wind load on ship}
Wind loads on ships, is an important parameter that need to calculated accurately,  because that is directly correlated with analysis of ship stability, maneuvering, station keeping and ship speed estimation. \cite{valvcic2016hybrid} developed a radial basis function neural network model by using elliptic Fourier features of closed contours and wind load data collected from wind load data of three types of ships, those are car carriers, container ships and offshore supply vessels. The trained neural network was employed for the prediction of non-dimensional wind load coefficients.             
\subsubsection{Shaft power prediction of large merchant ships}
\begin{figure}
\centering
\includegraphics[height=8cm]{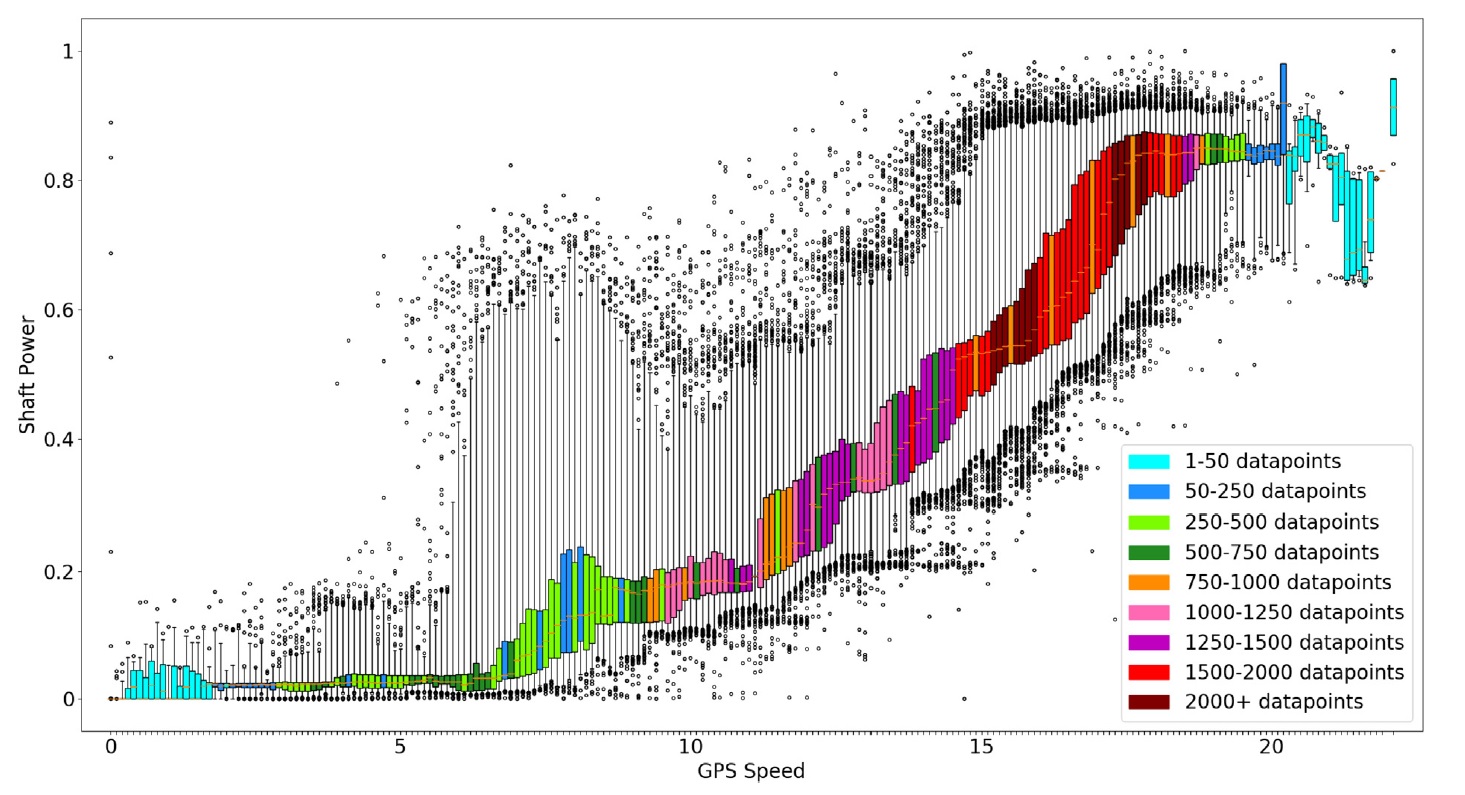}
\caption{Box-plot of shaft power vs GPS speed \citep{parkes2018physics}.} \label{fig:shaft power}
\end{figure}

\cite{parkes2018physics} used neural networks to predict shaft power of large merchant ships. The data required for training the neural network model was obtained from a data-set of 27 months of continuously monitored data   sampled in every 5 minutes from three vessels of same design(Sample data is shown in fig.\ref{fig:shaft power} reproduced from \cite{parkes2018physics}. The data consists of vessel movements recorded with varied geographic locations and weather conditions. The variables used in training of the neural network model are GPS ship speed, ship speed through water, true wind speed, apparent wind direction, draught, headings, trim and wave height. The above mentioned parameters/features were used as input to the neural network model and the output of the neural network is the shaft power. The shaft power is the product of shaft torque and angular velocity. From accurate measurement of shaft power the engine efficiency can be calculated.         
\subsubsection{Prediction of fuel consumption of ship main engine}
\cite{gkerekos2019machine} predicted the fuel consumption of main ship engine using different machine learning algorithms, those are support vector machines, random forest regressors, tree regressors, ensemble methods and artificial neural networks. \cite{gkerekos2019machine} mainly compared the predictive capability of above mentioned models in the prediction of the fuel consumption. Two different ship board data-sets were collected using two different strategies of data collection, noon reports and automated data logging and monitoring, were used for the model development. The ML models developed using different algorithms were found to be accurately predict the fuel consumption under different weather condition, load condition, sailing distance, drafts and speed. \cite{farag2020development} used ANN and multi-regression techniques to estimate the ship power and fuel consumption. The data used for the ML model development was obtained from \cite{farag2017decision}. The data-set consists of data from two sea voyages collected at different loading conditions. \cite{gkerekos2020novel} also used deep neural networks to develop a fuel consummation model and used that model in the route optimisation process. \cite{karagiannidis2021data} feed forward neural networks to 
predict the ship fuel consumption. The effect of data prepossessing on the model predictive accuracy was mainly analyzed. \cite{yuan2021prediction} used LSTM to model the real time fuel consumption of vessels and utilized the ML model to optimize the fuel consumption of the inland vessel using Reduced Space Searching Algorithm.

\begin{figure}
\centering
\includegraphics[height=12cm]{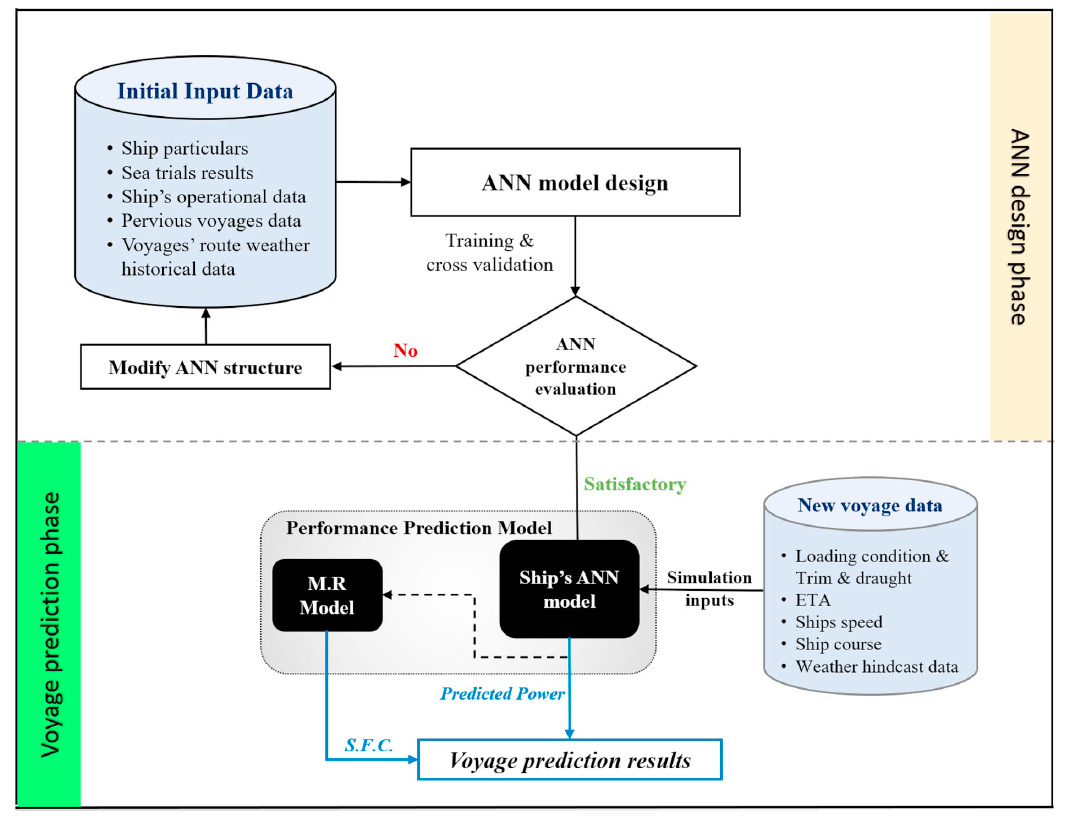}
\caption{ANN based model framework for fuel consumption prediction \citep{farag2020development}.} \label{fig:fuel_ann}
\end{figure}
\subsubsection{Ship collision avoidance}
\cite{gao2020ship} used generative adversarial network(GAN) to generate appropriate anthropomorphic collision avoidance decisions and bypass the process of ship collision risk
assessment based on the quantification of a series of functions. The LSTM cell was combined with GAN to improve capacity of memory and the current availability of the overall system. The data required for the development of the ML model was obtained from ship encounter azimuth map. The data-set has 12 types of ship encounter modes from automatic identification system(AIS) big data \citep{yang2021can}. The proposed ML model can be applied in intelligent collision avoidance, route planning and operational efficiency estimation.           

\subsection{Design and reliability analysis of breakwaters}
Breakwaters are constructed in ports and harbors to prevent an anchorage from the erosion from the harsh wave climate and long shore drift. The breakwaters the intensity of wave action in the ocean\citep{panduranga2021surface,kaligatla2021wave,vijay2020scattering}.  

\cite{kim2005neural} used artificial neural networks to design and reliability analysis rubble mound breakwater. The inputs to the neural networks are chosen by taking different combinations of the variables, $P$, $N$, $S_d$, $\zeta_m$, $cos\theta$, $h/H_s$, $SS$, $h/L_s$ and $T_s$. $h/H_s$ is the water depth parameter, $L_s$ is the period of significant wave, $h/H_s$ is the significant wave height, $P$ is the permeability of the breakwater, $N$ is the number of wave attack and the other parameters have their usual meaning as used in breakwater reliability analysis\citep{kim2005neural}. Based on different input combinations they used five neural networks to model the stability of the breakwater and the predictions of the neural networks are compared against conventional empirical model of \cite{van1990rock}. The stability model of \cite{van1990rock} has the from:
\begin{equation}
\begin{split}
N_{s}=6.2P^{0.18}(\frac{S_d}{\sqrt{N_w}})^{0.2} \frac{1}{\sqrt{\zeta_m}} 
\end{split}
\end{equation}
for $\zeta_m<\zeta_c$.

\cite{kim2014artificial} have used ANN to estimate the damage of breakwaters considering tidal level variations. They employed the wave height prediction neural network into a Montecarlo simulation. The ANN was used to predict the wave height in front of the breakwater. The inputs to the ANN are deep sea wave height and the other was the tidal level in front of the breakwater and the significant wave height is the output of the ANN.
\subsection{Detection of damaged mooring lines}
\cite{chung2020detection} used ANN to detect the damaged mooring line in tension leg platform through pattern analysis of floater responses. They used numerical simulation data(time series data of environment and floater responses) of charm3D for training and testing of the neural network. The environmental data are related to wave and wind, while the floater response data are related to the six degrees of freedom: surge,sway, heave, roll, pitch and yaw. They considered a ANN with five hidden layers in their modeling. The number of nodes in each layer as follows (16,100,300,500,300,100,9). 16 and 9 correspond to the first and last layer and other numbers indicate the number of neuron in respective hidden layers. Their ANN model was a classification network, whose job was to detect which mooring line is damaged by assigning an explicit label to it. \cite{aqdam2018health} used radial basis function neural network to detect the fault in the mooring line. The effects of uncertainties in the modeling (material, boundary, measurement uncertainties, hydrodynamic effects) were considered in developing the ML models.  
\subsection{Machine learning applied in propeller research}

\begin{figure}
\centering
\includegraphics[height=8cm]{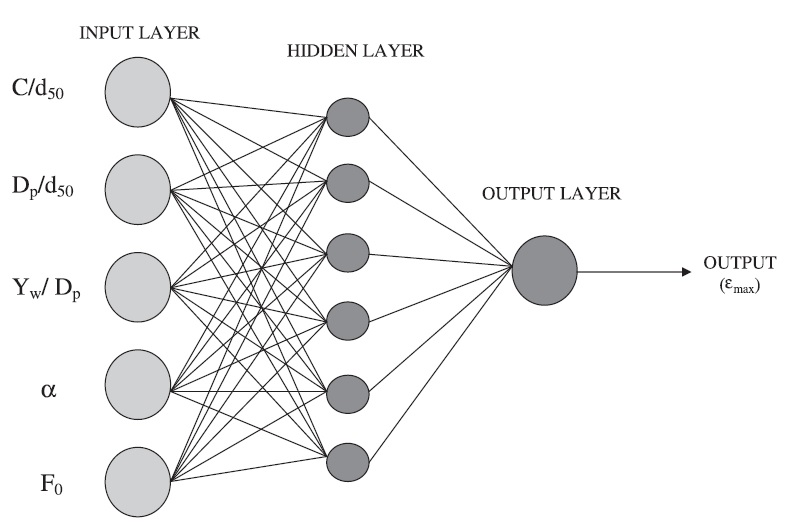}\\
\caption{Neural network used in modeling of the propeller induced erosion \citep{luo2014manoeuvring}.} \label{fig:propeller}
\end{figure}

Propellers are used in ships and submarines to create thrust to propel the vehicle \citep{prabhu2017fluid,kumar2017measurement,nandy2018heuristic} . The blades are designed in such a manner so that their rotational movement through the fluid generates a pressure difference between the two surfaces of blades. The propellers used in marine Engineering applications are mostly screw propellers with helical blades rotating on a propeller shaft. \cite{mahmoodi2019prediction} used gene expression programming(GEP) to evaluate hydrodynamic performance and cavitation volume of the marine propeller with various geometrical and physical conditions. CFD data of propeller thrust, torque and cavitation volume at different rake angle, pitch ratio, skew angle, advance velocity ratio and cavitation number are utilized in the development of the GEP model. The mathematical expressions are developed for torque, thrust and cavitation volume in terms of the physical and geometrical parameters.  \cite{shora2018using} used ANNs to predict the hydrodynamic performance and cavitation volume of propellers at different operating conditions. The data utilized in training and testing of the ANN model was obtained from CFD simulations of the flow past the propeller with varied geometrical and physical parameters. The input variables to the neural network are taken as rake angle, skew angle, pitch ratio, advanced ratio and cavitation number and the output variables are propeller thrust, propeller torque and cavitation volume. They generated 180 different data-sets by varying the input parameters. For different output, they considered different configuration of ANN for minimum mean squared error. They considered feed-forward and back-propagation ANNs in their simulations. Their ANN models very good prediction accuracy(greater than 0.99).
\cite{kim2021study} used convolutional neural networks to study the risk of propeller cavitation erosion. The CNN model was trained using various ship model test results of cavitation characteristics. Three types of CNN were used for the ML model development, those are VGG, GoogleNet and ResNet.
\cite{ryan2013determining} used ANN to model propeller induced erosion alongside quay walls in shallow water harbours and compared the predictive capability of the ANN by contrasting the results with other regression based models. The structure of the ANN used in modeling of the propeller induced erosion is shown in fig. \ref{fig:propeller}. The ANN has five inputs, one output and six hidden layers. The input parameters to the neural network are Clarence distance from propeller tip to bed, propeller diameter, distance to quay wall, rudder angle and densimetric Froude number and the output of the neural network is the depth of maximum scour at a particular time instant. The data required for the ANN model was obtained from deep rectangular GRP-lined plywood tank, using two open propellers.
\subsection{Damage detection of offshore platforms} 
Offshore platforms are large structures installed in deep seas with facilities drilling of well to explore natural gas and petroleum that lies in the seabed. These are damaged during their service life, because of complex marine environments and human factors. In order to ensure the safety of marine operations, the structural health monitoring such as, vibration based damage detection technique must be employed. The vibration based damage detection method can identify the presence, location severity of the damages of structures.           \cite{bao2021one} used one-dimensional convolutional networks to detect the damage sensitive features automatically from a offshore platform using raw strain response data. The CNN model was validated using numerical simulations of jacket-type offshore platform for random and regular wave excitation in different directions. Different damage locations and the noise effect were considered for finding the damage localization and damage severity. The feature extraction capability of the CNN was enhanced using the data pre-processing procedure based on convolution and deconvolution for noisy data. The CNN model developed was tested for three different cases: e.g. offshore platform subject to a sinusoidal excitation, a white noise extraction and a impulse excitation. Basically the model of \cite{bao2021one} is an extension of the work of \cite{abdeljaber2017real}, in which they had used CNN to detect the damage of a grandstand simulator at Qatar University.        
\subsection{Miscellaneous applications in the marine environment}
\subsubsection{Beach classification}
\cite{lopez2015morphological} applied SVM and ANN to classify nine different types of beaches, those are mainly micro-tidal sand and gravel beaches. The beach types are a) sand and gravel beaches, b) sand and gravel separated beaches, c) gravel and sand separated beaches, d) Gravel and sand beaches, e) pure gravel beaches, f) supported sand beaches, g) open sand beaches, h) bi-supported sand beaches, i) enclosed beaches. The ML model results are compared with results of discriminant analysis. The 14 variables used in the classification model development are: modality, $D_{50}$, $D_{10}$, $D_{90}$, source, breaking wave height perpendicular to the beach, frequency, direction associated with the wave height perpendicular to the beach, length profile, type profile, slope of the berm, distance to the source and Posidonia depth. The above mentioned terms has their usual definitions and is available in \cite{lopez2015morphological}. The ML models developed with the above 14 variables, were optimized with variation of neurons in the hidden layers and the SVM was modelled with different kernels, those are linear, polynomial, radial basis function and sigmoid.              
\begin{figure}
\centering
\includegraphics[height=12cm]{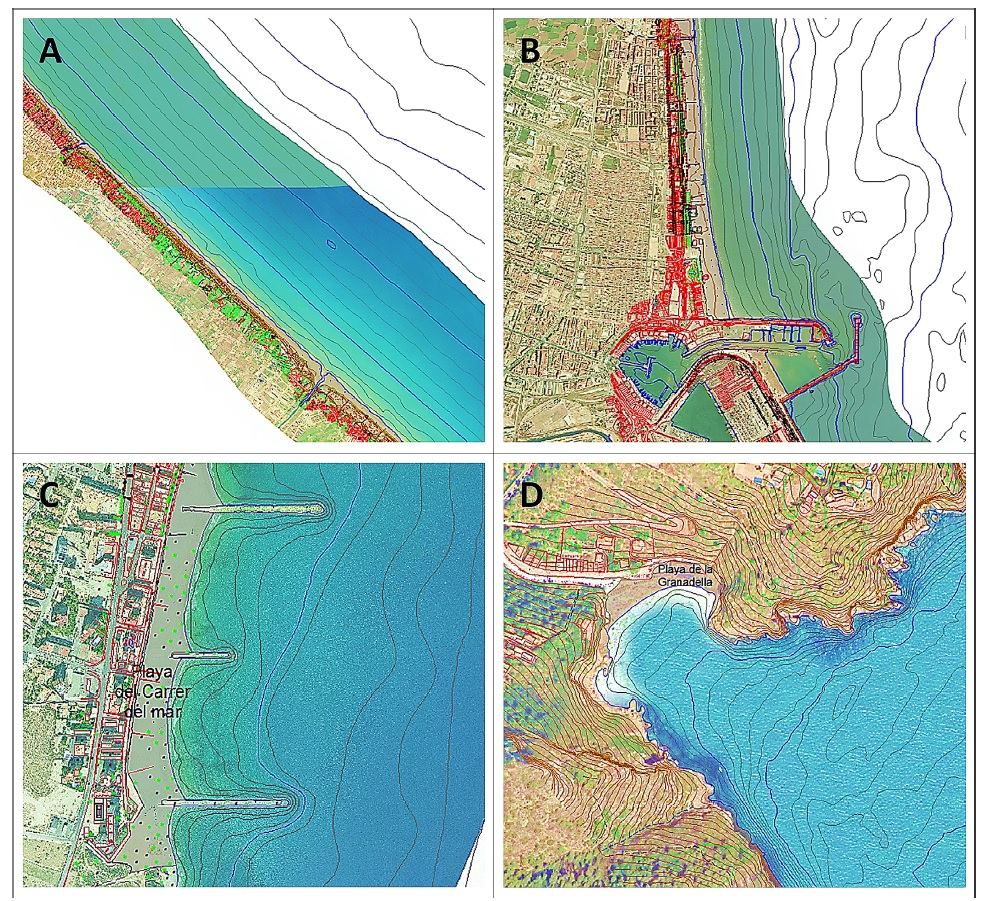}\\
\caption{Types of sandy beaches.(A) open sand beach.(B) supported sand beach.(C) bi-supported sand beach.(D) enclosed sand beach \cite{lopez2015morphological}.} \label{fig:beaches}
\end{figure}
\subsubsection{Condition monitoring of marine machinery system}
Condition based maintenance is an advanced data-driven method of machine maintenance, in which historical data collected by shipboard monitoring system is utilized by intelligence analysis tool to guide the planned maintenance. This makes the machine maintenance work, more scientific, systematic, and planned. \cite{tan2020comparative} used one class classification technology, that needs one class samples to train the model. Six different classifiers were used in the modeling of the condition monitoring system, those are one class support vector machine, support vector data description, Global k-nearest neighbors, Local outlier factor, Isolation forest, Angle-based outlier detection. In the development of ML models, the data-set of marine gas turbine propulsion system was used.  

\begin{figure}
\centering
\includegraphics[height=6cm]{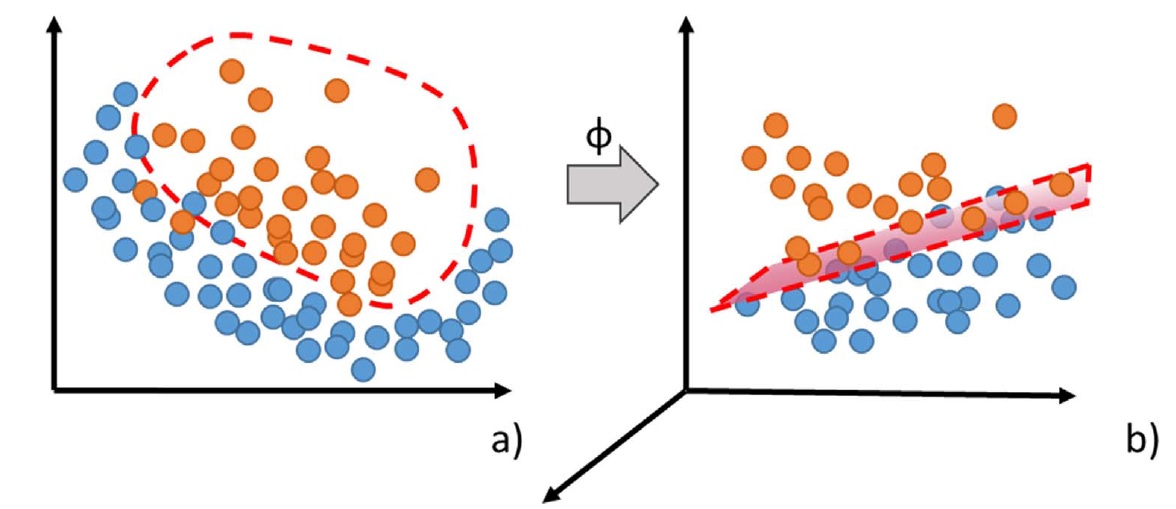}\\
\caption{Schematic of the support vector machine classification. A non-linear mapping was used to map the observations into a higher dimensional space \cite{pagoropoulos2017applying}.} \label{fig:svm}
\end{figure}
\subsubsection{Performance assessment
of shipping operations}
Energy efficient operations can lead to reduced fuel consumption and reduction in environment pollution. The improvements in the energy use efficiency can be achieved both by technical upgrades and through behavioural changes of the on board crew members. \cite{pagoropoulos2017applying} used multi-class support vector machines to identify the presence of energy efficient operations. The support vector machine utilized a radial basis function kernel, that facilitates the adaptive modeling of the interface between the classes and thus significantly improves classification performance as shown in fig.\ref{fig:svm}. The data required for developing the ML model was obtained through discussions with senior officers and technical superintends(mainly the positive and negative patterns of energy efficient operations were identified). The main source of data utilized in the model development was collected from noon reports \citep{poulsen2016logic} and based on the noon reports the energy consumption data were divided per consumer and covered the auxiliary machine parts used for generation of electricity, boilers, main engine, pumps and incinerators.
\subsubsection{Autonomous ship hull corrosion cleaning system}
\begin{figure}
\centering
\includegraphics[height=4cm]{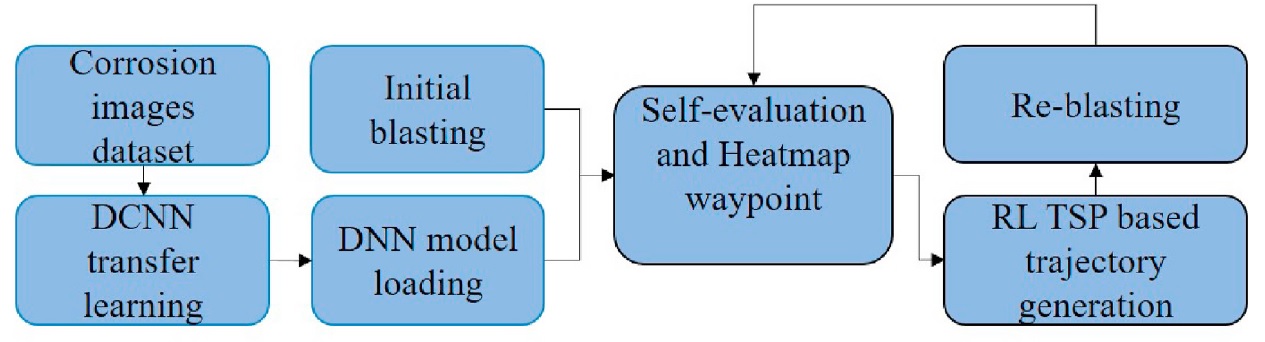}
\caption{The block diagram of close loop optimal automated water blasting \cite{le2021reinforcement}.} \label{fig:robot}
\end{figure}

The ships can be smoothly operated with cleaning of the hulls in shipyards. \cite{le2021reinforcement} used autonomous system based on reinforcement learning \citep{sutton2018reinforcement} to remove the corrosion of a ship by water blasting. The cleaning of ships by autonomous robotic systems ensure reduced consumption of water, time and energy in contrast to the manual cleaning. \cite{le2021reinforcement} developed a water blasting framework for a novel robot platform, which can navigate smoothly on a vertical plane and is designed with the adhesion mechanism of a permanent magnet. In order to ensure shortest travel distance and time to save resources used in the cleaning process, the complete way-point path planning is modeled as a classic travel salesman problem. The level of corroded areas after water-blasting was assessed by using deep convolutional neural networks. A detailed  discussion of the operation strategy of the robot is available in  \cite{le2021reinforcement}. The block diagram of the close loop optimal automated water blasting is shown in fig.\ref{fig:robot}.
\subsubsection{Wave energy forecasting}
Wave energy is a promising source of renewable energy. \cite{bento2021ocean} employed artificial neural networks to predict wave energy flux and other wave parameters. The The neural network model optimized using moth-flame optimization and the proposed model was assessed using 13 different data-sets collected from locations across the Atlantic,  Pacific coast and Gulf of Mexico. \cite{mousavi2021providing} used LSTM to forecast power generation of a wave energy converter using artificial neural networks. The effective forecast of wave energy will lead to reduced cost of investment in construction of the device and it is also essential for operation and management of electric power. They have used both experimental and numerical data for training and testing of the model. The experimental data was utilized from \cite{he2020coherence} and the numerical data was obtained from numerical simulations using Flow-3D software. In their study, a ML model was developed for a correlation between wave height and the generated electric power. \cite{vieira2021novel} developed a novel time efficient approach to calibrate VARANS-VOF models for simulation of wave interaction with porous structures using Artificial Neural Networks. These methods are useful for reducing time consumption in predicting the wave parameters using traditional fluid dynamics methods.     
\subsubsection{Prediction of wind and wave induced load effects on floating suspension bridges}
\begin{figure}
\centering
\includegraphics[height=4cm]{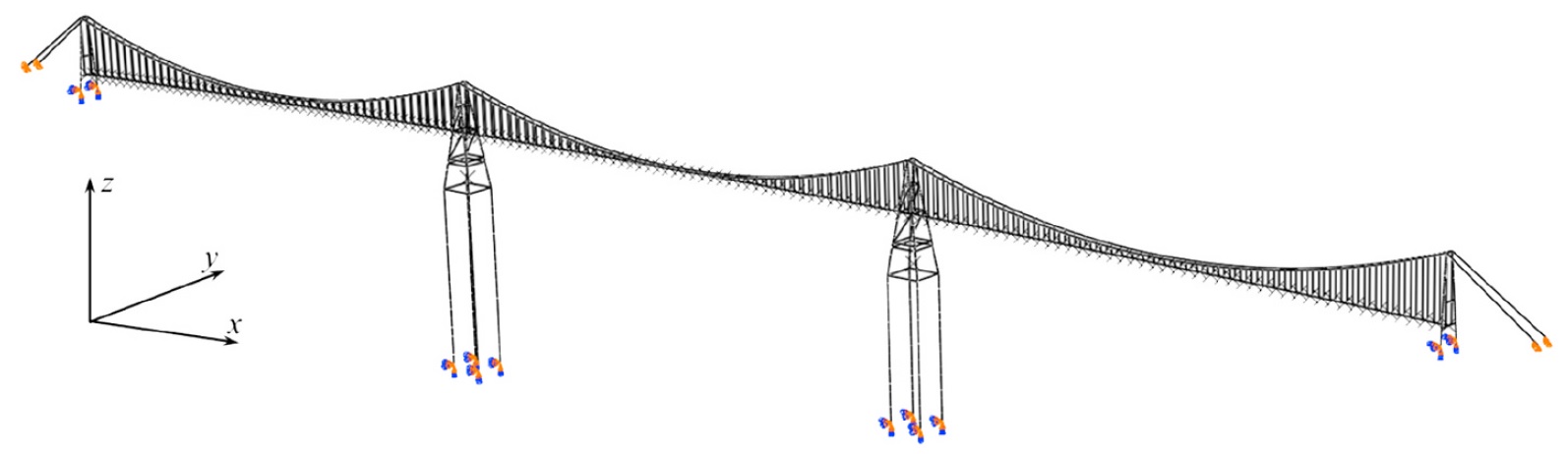}
\caption{A finite element model of 3-span suspension bridge with 2 floating pylons \citep{xu2020efficient}.} \label{fig:bridge}
\end{figure}

\cite{xu2020efficient} used ANN and SVM to predict long-term extreme load effects of floating suspension bridges. They used ML models as surrogate models, in conjunction with the Monte-carlo based methods, for the faster prediction of the loads. For their case study a 3-span suspension bridge with 2 floating pylons under combined wind and wave actions was used as shown in fig.\ref{fig:bridge}. In the new Monte-Carlo framework, the implicit limit-state function was replaced by the surrogate models based on ANN and SVM. It was noticed that, the ML based Monte-Carlo method require less computational effort and predicted more accurate results.     

\subsubsection{Tidal current prediction}
The traditional method of tidal current prediction employs computer applications with classical harmonic analysis. Those are computationally expensive for real time predictions. \cite{sarkar2018prediction} used Bayesian machine learning (Gaussian processes) for the prediction of tidal currents. With use of ML based techniques the uncertainties and complexity of the problem were enabled to represent in the modeling basis. The data used in the development of ML algorithm were collected from National Oceanic and Atmospheric Administration (NOAA). The location of tidal current observation sites are Southampton Shoal Channel, Old Port Tampa, Sunshine Sky Bridge and Martinez-AMORCO Pier. The data obtained from these sites were used for long-term predictions. The Gaussian process with periodic kernel function was found to be suitable for the modeling problem concerned, because of its harmonic nature. \cite{immas2021real} utilized deep learning models to develop tools for in situ prediction of ocean currents, those are, a Long Short-Term Memory (LSTM)
Recurrent Neural Network and a Transformer. The data utilized for model development were also obtained from NOAA. In addition to speed, they also predicted the direction of ocean currents. It was noticed that the predictions of Ocean currents are more accurate in contrast to the predictions of harmonic methods\citep{immas2021real}. \cite{sumangala2020coastal} used ANN to model the currents of Bay of Bengal. They mainly improved the velocity predictions using ANN model. 
\subsection{Application of ML in CFD}
The field of Naval Architecture, Ocean and Marine Engineering often employs CFD as a tool for modeling and prediction of flow past ships and underwater vehicles \citep{panda2021numerical,mitra2019effects,mitra2020experimental}, visualization of flow past propellers \citep{prabhu2017fluid}, analysis of wave induced load on offshore structures, simulations of waves and currents and simulations of wave energy converters \citep{mohapatra2021performance1, mohapatra2021performance2,mohapatra2020hydrodynamic} . The commercial software used in CFD analysis are ANSYS FLUENT \citep{fluent2011fluent}, STAR-CCM+ \citep{cd2017star} and SHIPFLOW \citep{larsson1989shipflow}. CFD is cheaper in comparison to experimental methods and can be applied for flow prediction in larger and complex domains, like modeling flow past larger ships and predicting complex flow past propellers. The basic building block of such CFD tools are turbulence models. In literature, ML techniques are either applied for faster flow prediction using surrogate models or developing turbulence closure models using large data-sets of direct numerical simulations(DNS) or experiments.  

In this section, we will provide a detailed overview of application of ML algorithms in CFD and turbulence modeling. The Naval Architects and CFD engineers working with different research problems may employ such advanced techniques for modeling and prediction of the complex oceanic flow fields.
\subsubsection{Fluid flow field prediction using reduced order models}
\cite{sekar2019fast} used deep neural networks for flow prediction over airfoils. They used a deep convolutional neural network for extraction of features from the different shapes of the airfoil and utilized those shape related features along with Reynolds number and angle of attack as inputs to the DNN model. The outputs to the DNN are pressure and velocity components across the airfoil. The data required for training the DNN model is obtained from CFD simulations. They predicted the flow field at much faster speed (150 times) in comparison to the traditional CFD methods and the results are as accurate as the traditional CFD predictions. \cite{hui2020fast} utilized CNN to predict pressure distribution around airfoils. The data-set library was formed using numerical simulation data from deformed airfoils. The airfoil was parametrized using the signed distance function(SDF).   
\begin{figure}
\centering
\includegraphics[height=8cm]{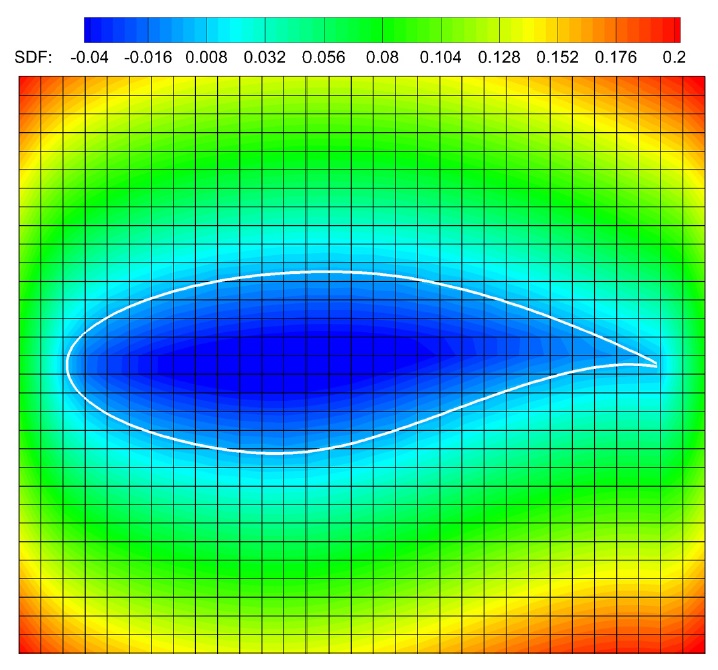}
\caption{The SDF representation of base airfoil \citep{hui2020fast}.} \label{fig:sdf}
\end{figure}
\cite{renganathan2020machine}   utilized DNN fro non-intrusive model order reduction of the parametric inviscid transonic flow past an airfoil. They preserved the accuracy of flow prediction at a significant lower computational cost. \cite{kong2021deep} used CNN to predict the velocity field for flow in in a scramjet isolator. The data required for training model was obtained from numerical simulations of flow at different Mach numbers and back pressures. The CNN has multiple reconstruction and feature extraction modules. A mapping relationship was established between the wall pressure on isolator and the velocity field on the isolator. \cite{kong2021data} used DL approach for super-resolution flow field reconstruction of a scramjet isolator. In contrast to the work of \cite{kong2021deep}, \cite{kong2021data} used experimental data-sets for the development of the DL model. They used both single path and multi-path network models based on CNNs for flow field reconstruction. It was noticed that the multi-path CNNs has better predictive accuracy in comparison to the single path CNNs. \cite{lee2021analysis} used CNN for predicting unsteady volume wake three-dimensional flow fields. The ML model was trained with past information of flow velocity and pressure. They also performed different analysis to find structural similarities among feature maps to reduce the number of feature maps containing redundant flow structures. Such reduction process can decrease the size of neural network without affecting the prediction performance. \cite{hasegawa2020machine} developed a reduced order model(ROM) for unsteady flow prediction by combining CNN and LSTM, which are trained in a sequential manner. The CNN model was trained using DNS data obtained from numerical simulations with 80 bluff bodies and tested on 20 bluff bodies. They also tested the ML-ROM model for unseen bluff bodies, and the predicted results were quite satisfactory, this shows the universality of ML-ROM based models. \cite{nakamura2021convolutional} modeled three dimensional complex flow using ROM. The ROM consists of a CNN and a LSTM. The function CNN was to map high dimensional flow fields into a low-dimensional latent space and the LSTM was used to predict temporal evolution of latent vectors.The data required for training the ROM was obtained from DNS. \cite{leer2021fast} utilized MLP in conjuction with  radial-logarithmic filter mask (RLF) for developing a universally applicable ML concept fast flow field estimation for various geometry types. The function of RLF is to provide information about the geometry in a compressed form for the MLP. They applied new concept for both internal and external flows such as airfoils and car shapes. The ML model was developed with data generated from CFD simulations for different geometries and the was tested for different unknown geometries.                               

\cite{sun2020surrogate} proposed a physics constrained deep learning (DL) approach for surrogate modeling of fluid flows. The surrogate model was developed without relying on CFD simulation data. They proposed a structured DNN to enforce the initial and boundary conditions and the Navier-Stokes (NS) equation were incorporated into the loss of the DNN in the training. \cite{kashefi2021point} proposed a novel deep learning (DL) approach (Point net architecture) for flow prediction in irregular geometries when the flow field is either a function of the size and shape of the bluff body or the shape of the domain. The grid vertices (spatial positions) of the mesh in the CFD domain were taken as the input to the DL model and corresponding flow parameters at those points were considered as output. The point net architecture learns the non-linear relationship between the inputs and outputs. The Point net model was trained for in-compressible laminar steady flow past a cylinder and for testing its generalizability, it was used for prediction of flow around multiple objects and airfoils. The ML model predictions are found to be satisfactory and also accurate.             
\begin{figure}
\centering
\includegraphics[height=4cm]{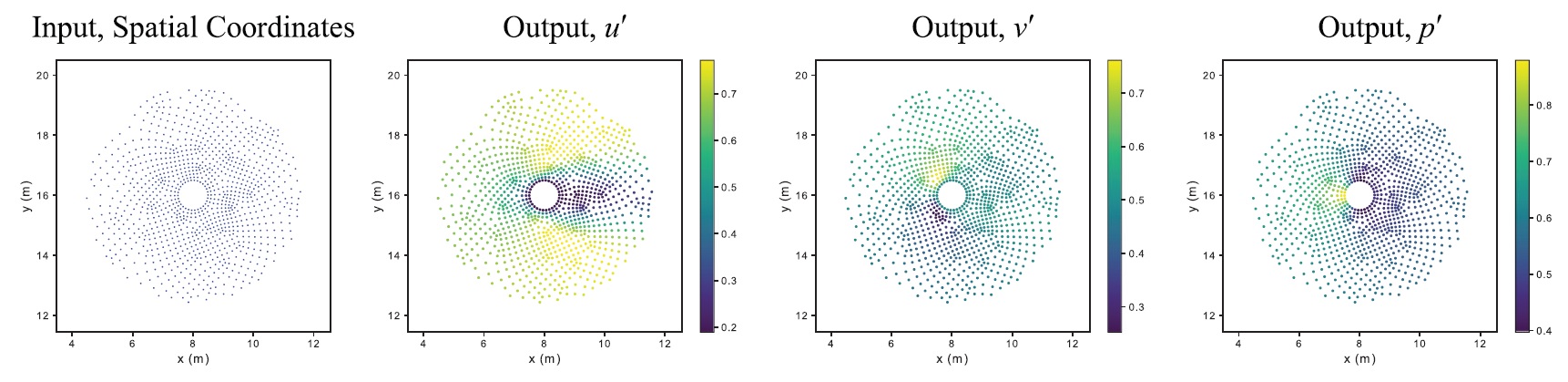}
\caption{Sample input and output data for the point-net \citep{kashefi2021point}.} \label{fig:pointcloud}
\end{figure}

\subsubsection{Turbulence modeling with ML}
Turbulent flows are a classification of fluid flows that are characterised by the manifestation of spatio-temporal chaos, hyper-sensitivity to perturbations, increased rates of diffusion, heat and mass transfer, etc. As has been observed turbulence is the norm and not the exception in nature \citep{moin1997tackling}. In engineering context the ability to predict the evolution of turbulence is of critical importance \citep{pope2000}. Because of the chaotic nature exact numerical prediction of turbulent flows is impracticable. Most engineering CFD studies rely on turbulence models to account for turbulence. The accuracy of CFD simulations are largely dependent on choice of turbulence models. 

The simplest class of turbulence models are the mixing length based models, also referred to as one equation models \citep{speziale1991analytical}. While these are simple and computationally inexpensive they require extensive input parameters for each case. Many of these parameters cannot be determined without reliable CFD simulations leading to a paradoxical situation. Because of this one equation models are not regularly used in engineering applications. The second category of turbulence models is the eddy viscosity based models that includes popular models like the $k-\epsilon$, $k-\omega$ models. These use the concept of an eddy viscosity to form a constitutive equation between the Reynolds stresses and the instantaneous mean gradients. Eddy viscosity based models are universal, robust and computationally inexpensive. But they have significant limitations in flows separation and re-attachment \citep{speziale1990,mishra2019estimating}, flows around inhomogeneities like walls \citep{pope2000}, flows with moderate to high degrees of anisotropy \citep{pope2000, mishra2019linear}, etc. The final category of turbulence models are Reynolds Stress Models, that utilize the Reynolds Stress Transport Equations to formulate individual transport equations for the components of the Reynolds stress tensor. These include the Launder-Reece-Rodi (LRR) model \citep{lrr}, the Speziale-Sarkar-Gatski (SSG) model \citep{ssg,ssmodel}, the Mishra-Girimaji model \citep{mishra2, mishra6}, etc. Reynolds stress models offer higher accuracy and robustness at a higher computational cost. However Reynolds stress models have limitations as well especially in flows with high influence of rotational effects \citep{mishra1}, significant streamline curvature \citep{mishra3}, etc. We observe that while there are many different turbulence models, they all have significant limitations that affect their accuracy, their applicability and their robustness. All the above mentioned models are developed by calibrating the model coefficients with respect to experimental or direct numerical simulation(DNS) data. With increase in computational facilities and data storage capacity and advancements in the ML techniques, current emphasis of turbulence researchers have been shifted towards development of turbulence models by using different ML algorithms \citep{jimenez2018machine} such as random forest, gradient boosting trees and deep neural networks. In this section, a detailed discussion turbulence models developed with ML approaches will be provided.

\cite{duraisamy2015new} proposed new methods in turbulence and transition modeling using neural networks and Gaussian processes. They developed improved functional forms using ML and applied those functional forms to predict the flow field. \cite{singh2017machine} used neural networks for developing model augmentations for the Spalart-Allmaras model. The model forms are constructed and incorporated into the CFD solver.  \cite{ling2016reynolds} used deep neural networks to develop a model for Reynolds stress anisotropy tensor by using high fidelity DNS data. They have proposed a new neural network (Tensor Basis Neural Network) that can accommodate the invariant properties (Galilean Invariance) into the modeling basis. The neural network structure was optimized using Bayesian optimization\cite{snoek2012practical}. A significant improvement in flow predictions were noticed when the ML model predictions are compared against the baseline RANS models. A data-driven Physics Informed Machine Learning(PIML) approach was proposed by \cite{wang2017physics}, in which the discrepancies in the RANS modeled Reynolds stresses were reconstructed. They used Random forests for modeling of the discrepancies with DNS data and the model developed was used to predict flow field for other flow cases, which were not used for model development. \cite{wu2018physics} proposed a systematic approach for choosing the input feature variables used in turbulence modeling. They considered strain rate tensor, rotation rate tensor, pressure gradient, TKE gradient, Wall distance based Reynolds number, turbulence intensity and ratio of turbulent time-scale to mean strain time-scale as the input features to model the discrepancy between RANS modeled and true Reynolds stresses. They used gradient of the flow features in place of the actual value to ensure Galilean invariance \citep{pope2000}. Finally the predictive capability of the ML model was tested against square duct and periodic hill flows. 
\cite{kaandorp2020data} proposed Tensor Basis Random Forest(TBRF), a novel machine learning algorithm to predict the Reynolds stress anisotropy tensor(RSAT).  The use of tensor basis ensures the Galilean invariance in the prediction of the RSAT. The TBRF was trained with various flow cases of DNS/LES data and was tested for unseen flows. The predicted values of the RSAT was finally employed in a RANS solver for flow prediction in unseen flows. \cite{zhu2021turbulence} developed surrogate turbulence model for prediction of flow past airfoils at high Reynolds numbers. Rather than using DNS or LES data, they utilized results from numerical simulations of Spallart-Allmaras (SA) model for training the DNN. The model was trained with six different six different free stream conditions for NACA0012 airfoil.     

\cite{panda2021modelling} have proposed a data-driven model for the pressure strain correlation of turbulence\citep{panda2018representation,panda2020review} using Reynolds stress anisotropy, dissipation, turbulence kinetic energy and the strain rate as the input to the DNN . They used DNS data of flow in channels at different friction Reynolds numbers for the development of the DNN based model and the model was tested for flows in different Reynolds numbers and also for a fully unknown test case of Couette flow in channels. The model predictions was also contrasted against other established pressure strain correlation models. \cite{panda2021evaluation} compared the predictive capability of random forest, gradient boosted tree and DNN in Reynolds stress transport modeling. They mainly considered the modeling of the pressure strain correlation of turbulence. Using Bayesian optimization they recommended the optimal hyper-parameters of the DNN.

\cite{wang2018investigations} proposed ML based subgrid-scale models(SGS) using DNN and RF for LES. They considered 30 flow variables as inputs to the ML model and analysed the feature importance of input variables and found that the filtered velocity and second derivative of the input velocity has larger influence on the output variable. The newly proposed ANN based SGS model has a correlation coefficient of 0.7. The proposed ANN based SGS model was found to be accurate than the Smagorinsky model and the
dynamic Smagorinsky model for flow predictions in isotropic turbulence. \cite{yuan2020deconvolutional} used a deconvolutional artificial neural network(DANN) for modeling SGS. The input features for the DANN are the filtered velocities at different spatial points. It was observed that the DANN model predicted the SGS stress more accurately than the approximate deconvolution method and the velocity gradient model with a corrrelation coefficient of $0.99$. \cite{xie2020artificial} developed ANN based algebraic models for the SGS in LES of turbulence at the Taylor Reynolds number ranging from 180 to 250. They had constructed the coefficients of the non-linear algebraic model using the ANN. It was shown that the ANN based non-linear algebraic model predicted the SGS stress more accurately in a priori tests.

\section{Challenges and Priority Research Directions}
Machine learning applications are having significant successes and impact across different problems in marine engineering, ocean engineering and naval architecture. Nonetheless we need to ensure a higher degree of trust in ML models, enable adequate verification and validation (V\&V) of data driven models, and utilize the strengths of data driven algorithms together with the decades of purely physics based understanding that has been developed in these fields \citep{hu2020physics}. 

An important step is to include physics and domain knowledge in machine learning models \citep{baker2019workshop}. This can be done at various levels for example the choice of the model algorithm and its hyperparameters, the features that are inputted to the algorithm, or directly appending physics constraints in the loss functions and optimization. As an illustration we can enforce mass and momentum conservation as constraints by appending additional losses that penalize the violation of these constraints in the loss functions of the turbulence models generated by deep neural networks. This would ensure more physical data driven models. This would also reduce the space of functions that the optimizer has to query over and lead to better models that require less training data. 

Another important step is generating measures of interpretability and explainability from ML models \citep{zaman2020scientific}. As an illustration in physics based models that are commonplace in marine and ocean engineering every term and each expression has physical meaning and represents different physics based interactions and processes. Machine learning models do not allow such understanding and rationalization. This lack of transparency leads to an understandable lack of trust in ML models from scientists and engineers working in marine and ocean applications. This also obviates any model validation. Verification and validation are essential steps to be executed before the deployment of any physics based models in marine and ocean engineering and should be so for data driven models as well. Model verification for ML models can be and is carried out using test datasets to estimate generalization error. For model validation we need to be able to unearth the model's reasoning for its predictions. Due to the black box nature of algorithms like deep neural networks, this is not possible yet. Hence there is a critical need to interpret and explain the reasoning of trained ML models before they can start to replace traditional physics and empiricism based models. 

A final need is to ensure robustness in the performance of ML models \citep{hegde2020quantifying}. Traditional models in marine and ocean applications have been based on physics. Physical laws such as symmetries, conservation of mass, momentum and species, etc are universal and extend to all ranges of parameter space. But machine learning models are restricted to the range where training data was utilized for their optimization. In regions of feature space far from the training data ML models make extremely poor predictions. For their general application there is a need to guarantee robustness in model performance.

\section{Concluding remarks}
In this article, we have provided a detailed review of application of machine learning algorithms in ocean engineering, naval architecture and marine engineering applications. The different machine learning algorithms are discussed in detail. The ML applications in the marine environment were classified into several categories such as wave forecasting, AUV operation and control, ship research, design and reliability analysis of breakwaters, applications in propeller research etc. The features used in modeling different marine processes and parameters were discussed in detail. The source of data utilized in model development are presented. The features used as inputs to the ML models are discussed in detail. Different algorithms used in optimization of the ML models were also discussed. A detailed overview of application of ML in CFD and turbulence modeling were also presented. Based on this comprehensive review and  analysis we point out future directions of research that may be fruitful for the application of ML to ocean and marine engineering as well as problems in naval architecture. This review article will provide an avenue for marine engineers and naval architects to learn the basics of ML models and its applications in the ocean engineering, naval architecture and marine engineering applications.    





\newpage
\bibliographystyle{elsarticle-harv}\biboptions{authoryear}
\bibliography{asme2e.bib}







\end{document}